\documentclass[conference]{IEEEtran}
\IEEEoverridecommandlockouts
\usepackage{cite}
\usepackage{amsmath,amssymb,amsfonts}
\usepackage{algorithmic}
\usepackage{graphicx}
\usepackage{textcomp}
\usepackage{xcolor}
\def\BibTeX{{\rm B\kern-.05em{\sc i\kern-.025em b}\kern-.08em
    T\kern-.1667em\lower.7ex\hbox{E}\kern-.125emX}}
    
\usepackage{booktabs}       
\usepackage{nicefrac}       
\usepackage{microtype}      
\usepackage{subcaption}
\usepackage{multirow}
\usepackage{hyperref}

\usepackage{flushend}
\begin{document}

\title{Semi-Supervised Graph Learning Meets
Dimensionality Reduction\\
\thanks{We acknowledge support for this work from two NSF grants, one NIH grant, and three DOE grants. Furthermore, we acknowledge support by the Fullbright program offered through the Thailand-United States Educational Foundation.}
}


\author{\IEEEauthorblockN{1\textsuperscript{st} Alex Morehead}
\IEEEauthorblockA{\textit{Department of EECS}\\
\textit{University of Missouri}\\
acmwhb@missouri.edu}
\and
\IEEEauthorblockN{1\textsuperscript{st} Watchanan Chantapakul}
\IEEEauthorblockA{\textit{Department of EECS}\\
\textit{University of Missouri}\\
w.chantapakul@missouri.edu}
\and
\IEEEauthorblockN{2\textsuperscript{nd} Jianlin Cheng}
\IEEEauthorblockA{\textit{Department of EECS}\\
\textit{University of Missouri}\\
chengji@missouri.edu}
}

\maketitle

\begin{abstract}
Semi-supervised learning (SSL) has recently received increased attention from machine learning researchers. By enabling effective propagation of known labels in graph-based deep learning (GDL) algorithms, SSL is poised to become an increasingly used technique in GDL in the coming years. However, there are currently few explorations in the graph-based SSL literature on exploiting classical dimensionality reduction techniques for improved label propagation. In this work, we investigate the use of dimensionality reduction techniques such as PCA, t-SNE, and UMAP to see their effect on the performance of graph neural networks (GNNs) designed for semi-supervised propagation of node labels. Our study makes use of benchmark semi-supervised GDL datasets such as the Cora and Citeseer datasets to allow meaningful comparisons of the representations learned by each algorithm when paired with a dimensionality reduction technique. Our comprehensive benchmarks and clustering visualizations quantitatively and qualitatively demonstrate that, under certain conditions, employing $\textit{a priori}$ and $\textit{a posteriori}$ dimensionality reduction to GNN inputs and outputs, respectively, can simultaneously improve the effectiveness of semi-supervised node label propagation and node clustering. Our source code is freely available on \href{https://github.com/amorehead/SSL-With-DR-And-GNNs}{GitHub}.
\end{abstract}


\section{Introduction}
Semi-supervised learning (SSL), a sub-discipline of unsupervised learning, is focused on transferring information about labeled class samples to unlabeled samples \cite{van2020survey}. Recently, semi-supervised learning in the context of graph neural networks (GNNs) has received increased attention by researchers for tasks such as node \cite{xu2020label} and graph \cite{kang2021structured} classification. Under the cluster assumption \cite{van2020survey}, such efforts, in many common formulations, aim to increase the robustness and accuracy of class clusterings produced on a given dataset through effective propagation of node label information. Concurrently, dimensionality reduction techniques have been widely used throughout several domains in machine learning and other scientific disciplines \cite{reddy2020analysis}. Nonetheless, few works have explored the quantitative and qualitative effects of $\textit{a priori}$ and $\textit{a posteriori}$ dimensionality reduction algorithms on the clusterings produced by GNNs, where the dimensionality of each graph's node representations is reduced before and after a network's forward pass, respectively. In this work, we explore such effects by investigating the behavior of GNNs for transductive tasks such as semi-supervised node classification on the Cora \cite{10.5555/3041838.3041901} and Citeseer \cite{sen2008collective} datasets.

\section{Related Work}
Few works have explicitly considered leveraging the connection between semi-supervised learning, graph-based learning, and graph-regularized dimensionality reduction to enrich the clustering and classification outputs of GNNs. One of the earliest works on SSL can be found in \cite{nigam2000text} where authors train Gaussian mixture models to classify text based on labeled and unlabeled documents. Another early work is that of \cite{chapelle2006continuation}, where continuation, a global (non-convex) optimization scheme, is introduced to train semi-supervised support vector machines for image classification across a variety of image datasets. More recently, \cite{berthelot2019mixmatch} introduce MixMatch, a deep learning-based approach to SSL on image data. Simultaneously, graph-based deep learning methods have seen a rise in popularity and adoption over the last several years. Notably, works such as those of \cite{DBLP:journals/corr/KipfW16} and \cite{yang2016revisiting} demonstrate the effectiveness of GNNs for label propagation in a semi-supervised manner.

In an adjacent way, researchers have developed new methods for dimensionality reduction on manifold datasets \cite{van2009dimensionality}. As graphs represent one of many kinds of manifolds, researchers have since applied dimensionality reduction techniques to graph data \cite{yan2006graph, zhang2012graph}. Similarly, \cite{mao2015dimensionality} created a method for learning from graph structures to perform dimensionality reduction in a principled manner. Interestingly, \cite{zhao2020connecting} mathematically relate the concepts of graph-regularized PCA to recent developments in graph-based neural networks. Nonetheless, no works have directly and intentionally pursued a comprehensive investigation into the effect of dimensionality reduction algorithms on the inputs and outputs of GNNs. As such, in the proceeding sections, we describe our efforts into elucidating and exploring this idea further, ultimately to determine whether any notable observations or principles may be concretized.

\section{Methods}
\subsection{Semi-supervised Learning}

Our study employs SSL through the way in which we backpropagate loss through each deep learning model trained. Namely, we mask loss gradients corresponding to unlabeled nodes, where unlabeled nodes are those encountered during training that were not originally designated to serve in our training dataset. The remainder of our networks' operations follow conventional deep learning practices \cite{goodfellow2016deep}.

\subsection{Graph Neural Networks}

To date, a plethora of GNNs has been developed for SSL on transductive graph datasets. In this work, we explore four deep learning models as methods to perform representation learning or graph message-passing on the Cora and Citeseer datasets. Such models include a multi-layer perceptron (MLP), the Graph Convolutional Network (GCN) \cite{DBLP:journals/corr/KipfW16}, the Graph Attention Network (GAT) \cite{velivckovic2017graph}, and the Graph Convolution method (GraphConv) of \cite{morris2019weisfeiler}. First, we let $\phi$ and $\psi$ represent distinct MLPs, $\bigoplus$ denote tensor addition, $c_{ij}$ be the edge weight between nodes $i$ and $j$, and $a(\mathbf{x}_{i}, \mathbf{x}_{j})$ serve as the attention score between nodes $i$ and $j$. Subsequently, the equations

\begin{equation}
    \mathbf{h}_{i} = \phi \left( \mathbf{x}_{i} \right),
    \label{eq:mlp_equation}
\end{equation}

\begin{equation}
    \mathbf{h}_{i} = \phi \left( \mathbf{x}_{i}, \bigoplus_{j \in N_{i}} c_{ij} \psi(\mathbf{x}_{j}) \right),
    \label{eq:gcn_equation}
\end{equation}

\begin{equation}
    \mathbf{h}_{i} = \phi \left( \mathbf{x}_{i}, \bigoplus_{j \in N_{i}} a(\mathbf{x}_{i}, \mathbf{x}_{j}) \psi(\mathbf{x}_{j}) \right), \ \mathrm{and}
    \label{eq:gat_equation}
\end{equation}

\begin{equation}
    \mathbf{h}_{i} = \phi_{1} \left( \mathbf{x}_{i} \right) + \phi_{2} \left( \bigoplus_{j \in N_{i}} c_{ij} \mathbf{x}_{j} \right)
    \label{eq:graphconv_equation}
\end{equation}

describe how our MLP, GCN, GAT, and GraphConv models, respectively, update with each network layer the current representation $\mathbf{x}_{i}$ for node $i$ to the updated node representation $\mathbf{h}_{i}$. For node classification, after applying a variable number of network layers to our input graphs, we then prepare our graphs' learned node representations for classification by reducing their dimensionality down to the respective number of node classes in a given dataset using a final chosen network layer.

\subsection{Dimensionality Reduction}
When one builds a machine learning model by letting it learn from the data it is provided, one may be convinced that utilizing all available features will be beneficial to the model's performance and robustness. However, if such features are in a high-dimensional space (e.g., $\vec{x} \in \mathbb{R}^{300}$), this scenario can vividly illustrate the well-known curse of dimensionality that affects how each model behaves. As such, some models may not perform well on higher dimensional data even when they perform well in a lower dimensional space. Fortunately, there are various techniques that we can employ to mitigate the curse of dimensionality. Feature selection methods such as Recursive Feature Elimination (RFE) \cite{chen2007enhanced} can be used to do so. Alternatively, one could apply dimensionality reduction techniques as well. These methods directly transform data from a high-dimensional space into a low-dimensional space (e.g., $\mathbb{R}^{3}$). The following subsections describe the four dimensionality reduction algorithms we applied to the inputs or outputs of our chosen GNNs, respectively, to simultaneously reduce input data dimensionality for enhanced semi-supervised node classification performance and to perform robust node-based clustering by visualizing the node representations learned by each semi-supervised GNN after reducing their dimensionality to $\mathbb{R}^{2}$.

\subsubsection{Principle Component Analysis}
Principle Component Analysis (PCA)---or Karhunen-Loeve transformation---is one of the most widely used dimensionality reduction techniques since it is an unsupervised algorithm. Invented in 1901 by Pearson \cite{doi:10.1080/14786440109462720}, PCA operates as follows. Given a vector $\vec{x} \in \mathbb{R}^d$, we define an orthonormal matrix $\mathbf{A} \in \mathbb{R}^{d \times d}$. The matrix-vector multiplication between $\mathbf{A}$ and $\vec{x}$ is then given by
\begin{equation}
    \vec{y} = \mathbf{A}^\mathsf{T} \vec{x},
    \label{eq:pca:transform}
\end{equation}
a transformation that, once applied to $\vec{x}$, transforms $\vec{x}$ into $\vec{y}$ in the corresponding new coordinate space. The core idea behind PCA is to preserve the variance of the original data being transformed. Hence, a covariance matrix $\mathbf{\Sigma}$ that represents the covariance between pairs of dimensions plays an important role in PCA. In particular, the eigendecomposition of $\mathbf{\Sigma}$ yields
\begin{equation}
    \mathbf{\Sigma}\mathbf{A} = \mathbf{A}\mathbf{\Lambda},
    \label{eq:pca:eig}
\end{equation}
where $\mathbf{A}$ and $\mathbf{\Lambda}$ are the corresponding eigenvector and eigenvalue matrices, respectively. In the context of PCA, we often keep only the $k$ eigenvectors (denoted as $\mathbf{A}_k$) that correspond to the $k$-largest eigenvalues in $\mathbf{\Lambda}$. In using such results, we get the transformed vector
\begin{equation}
    \underset{k\times 1}{\vec{y}} = \underset{k\times k}{\mathbf{A}_k^\mathsf{T}}\ \ \underset{d\times 1}{\vec{x}}.
\end{equation}
Since we normally choose $k << d$, this results in a vector in a much lower dimensional space $\vec{y} \in \mathbb{R}^k$. Furthermore, one could set $k = 2$ or $k = 3$. Doing so enables one to plot and visualize the transformed data in $\mathbb{R}^{2}$ or $\mathbb{R}^{3}$ as such visualizations may be useful for gaining an enhanced intuition regarding the original data's clustering characteristics.

\subsubsection{t-distributed Stochastic Neighbor Embedding}
The t-distributed Stochastic Neighbor Embedding (t-SNE) algorithm \cite{JMLR:v9:vandermaaten08a} has seen wide use by the machine learning community over the last decade \cite{van2008visualizing} for tasks such as visualization of the representations learned by convolutional neural networks. The key equations defining its usage are as follows:

\begin{equation}
    p_{ij} = \frac{\exp(-|| x_i - x_j ||^2) / 2\sigma^2}{\sum_{k \neq i}\exp(-|| x_i - x_k ||^2) / 2\sigma^2},
    \label{eq:tsne:p_ij}
\end{equation}

\begin{equation}
    q_{ij} = \frac{(1 + ||y_i + y_j||^2)^{-1}}{\sum_{k \neq l} (1 + || y_k - y_l ||^2)^{-1}}
    \label{eq:tsne:q_ij}.
\end{equation}
Employing a perplexity of 40 for our datasets, the primary objective of t-SNE is to minimize the Kullback-Leibler (KL) divergence \cite{van2014renyi} based on the high-dimensional pairwise similarity $p_{ij}$ between point $i$ and point $j$ and the low-dimensional pairwise similarity $q_{ij}$ between point $i$ and point $j$. Here, $x_i$ denotes a high-dimensional point, and $y_i$ denotes a low-dimensional point. It can be written in a mathematical form as follows:
\begin{equation}
    \underset{y_1, y_2, ..., y_n}{\min}\sum_{i=1}^{n} \mathrm{KL}(P_i || Q_i)
    = \underset{y_1, y_2, ..., y_n}{\min} \sum_{i=1}^{n} \sum_{j=1}^{n} p_{ij} \log \frac{p_{ij}}{q_{ij}}
    \label{eq:tsne:full}
\end{equation}

\subsubsection{Uniform Manifold Approximation and Projection}
Since its release in 2018 \cite{mcinnes2018umap}, the Uniform Manifold Approximation and Projection (UMAP) algorithm has been explored with great interest by machine learning researchers as a replacement to t-SNE \cite{becht2019dimensionality}. It is established upon a representation of a weighted graph. UMAP aims to maintain the local structure and the global structure at the same time after the data is projected down to $k$-dimensional space through minimizing the cost function which is defined as
\begin{equation}
    C_\mathrm{UMAP} = \sum_{i \neq j}( \underbrace{v_{ij} \log \frac{v_{ij}}{w_{ij}}}_{\scriptstyle\text{being neighbors}}
    + \underbrace{(1 - v_{ij}) \log \frac{1 - v_{ij}}{1 - w_{ij}}}_{\scriptstyle\text{not being neighbors}} ),
    \label{eq:umap:cost}
\end{equation}
where $v_{ij}$ and $w_{ij}$ are the weights between node $i$ and node $j$ in the high dimensional space and in the low dimensional space, respectively.

\begin{figure}[t!]
    \centering
    \begin{subfigure}{0.48\textwidth}
    \includegraphics[width=\linewidth]{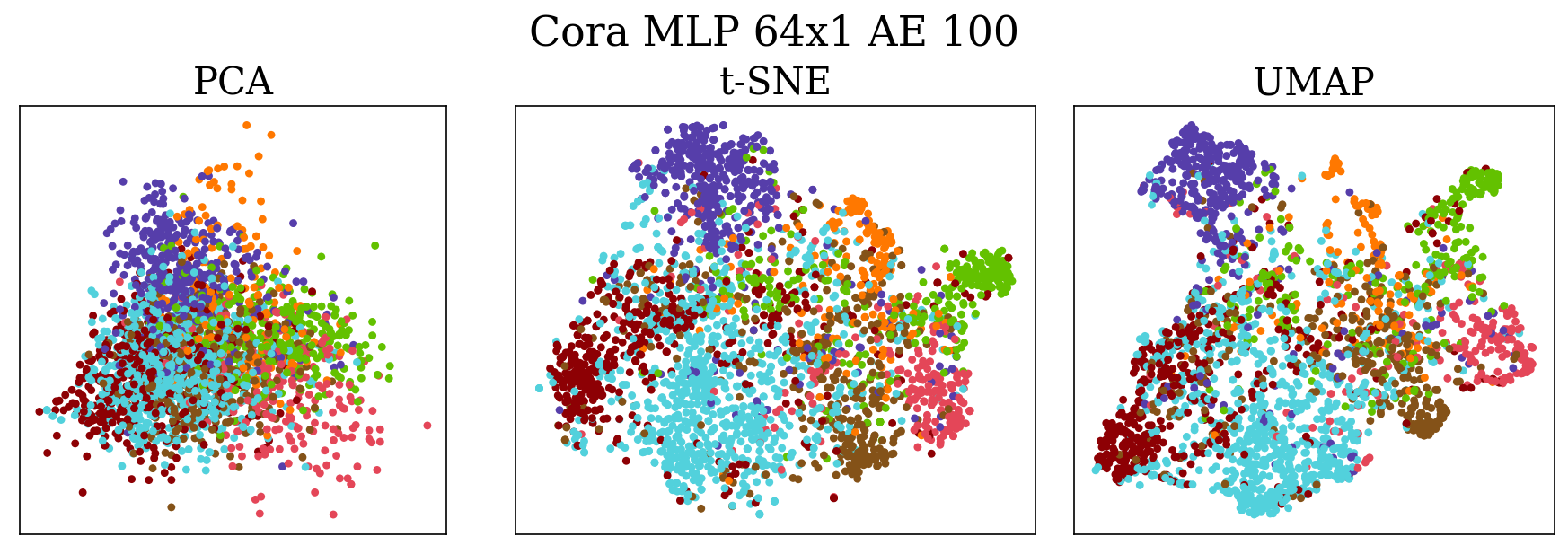}
    \includegraphics[width=\linewidth]{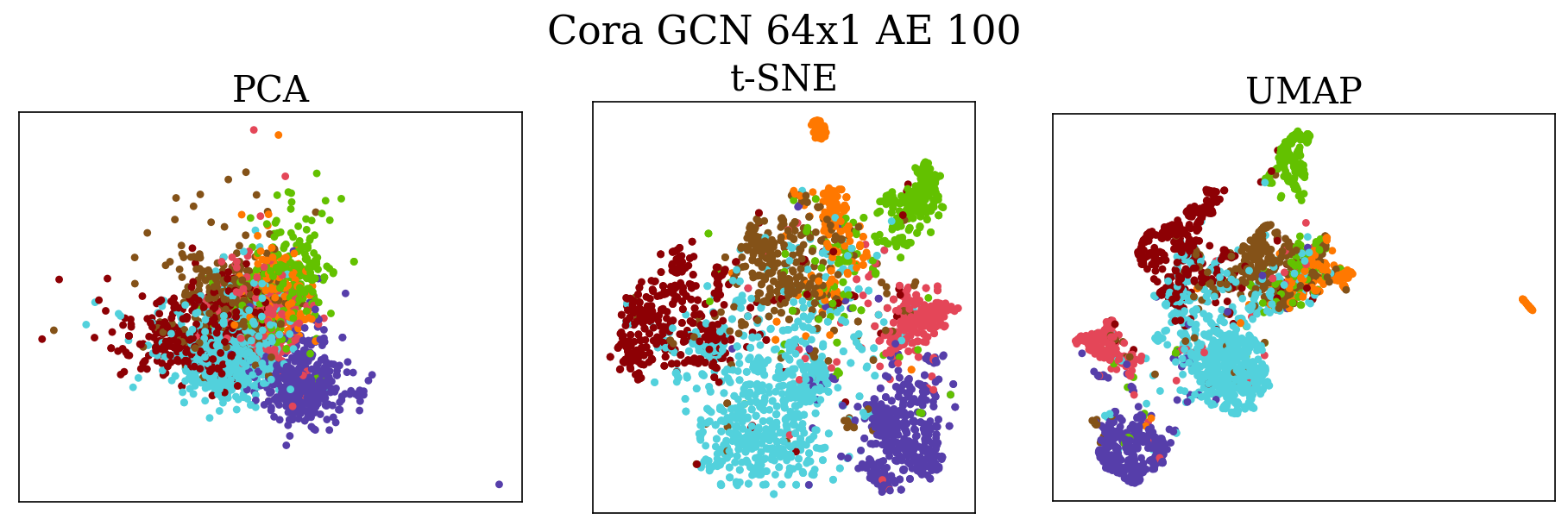}
    \caption{MLP and GCN clustering results with an AE applied $\textit{a priori}$.}
    \label{fig:cora_ap_ae_for_mlp_gcn}
    \end{subfigure}
    \begin{subfigure}{0.48\textwidth}
    \includegraphics[width=\linewidth]{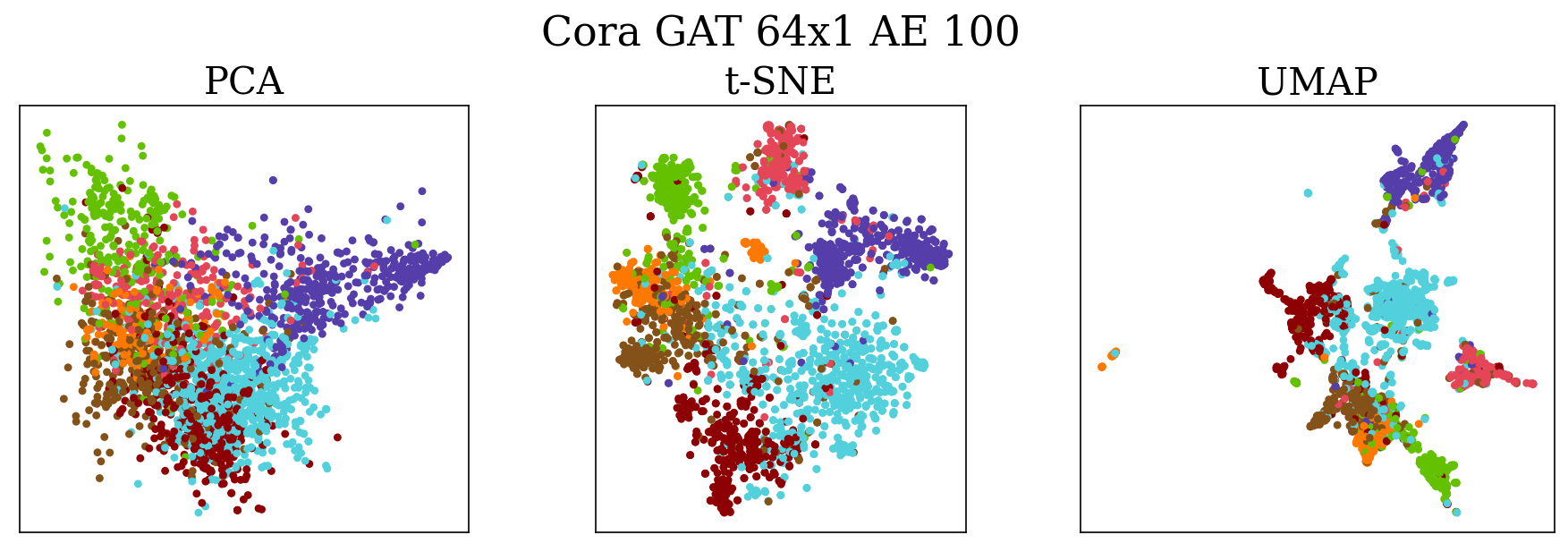}
    \includegraphics[width=\linewidth]{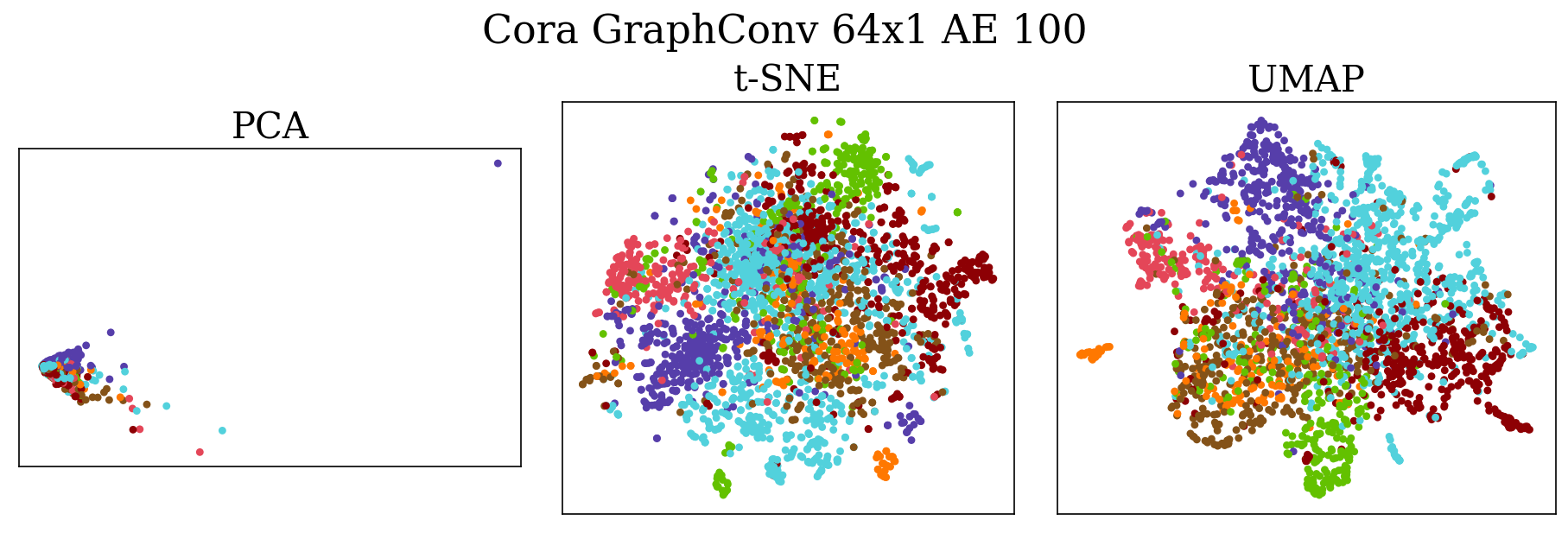}
    \caption{GAT and GraphConv clustering results with an AE applied $\textit{a priori}$.}
    \label{fig:cora_ap_ae_for_gat_graphconv}
    \end{subfigure}
    \caption{All models' clustering results on the Cora dataset with an AE applied $\textit{a priori}$.}
  \label{fig:ap_ae_mlp_gcn_gat_graphconv_on_cora}
\end{figure}

\subsubsection{Autoencoder}

\begin{figure}[hb]
  \centering
  \includegraphics[width=0.6\linewidth]{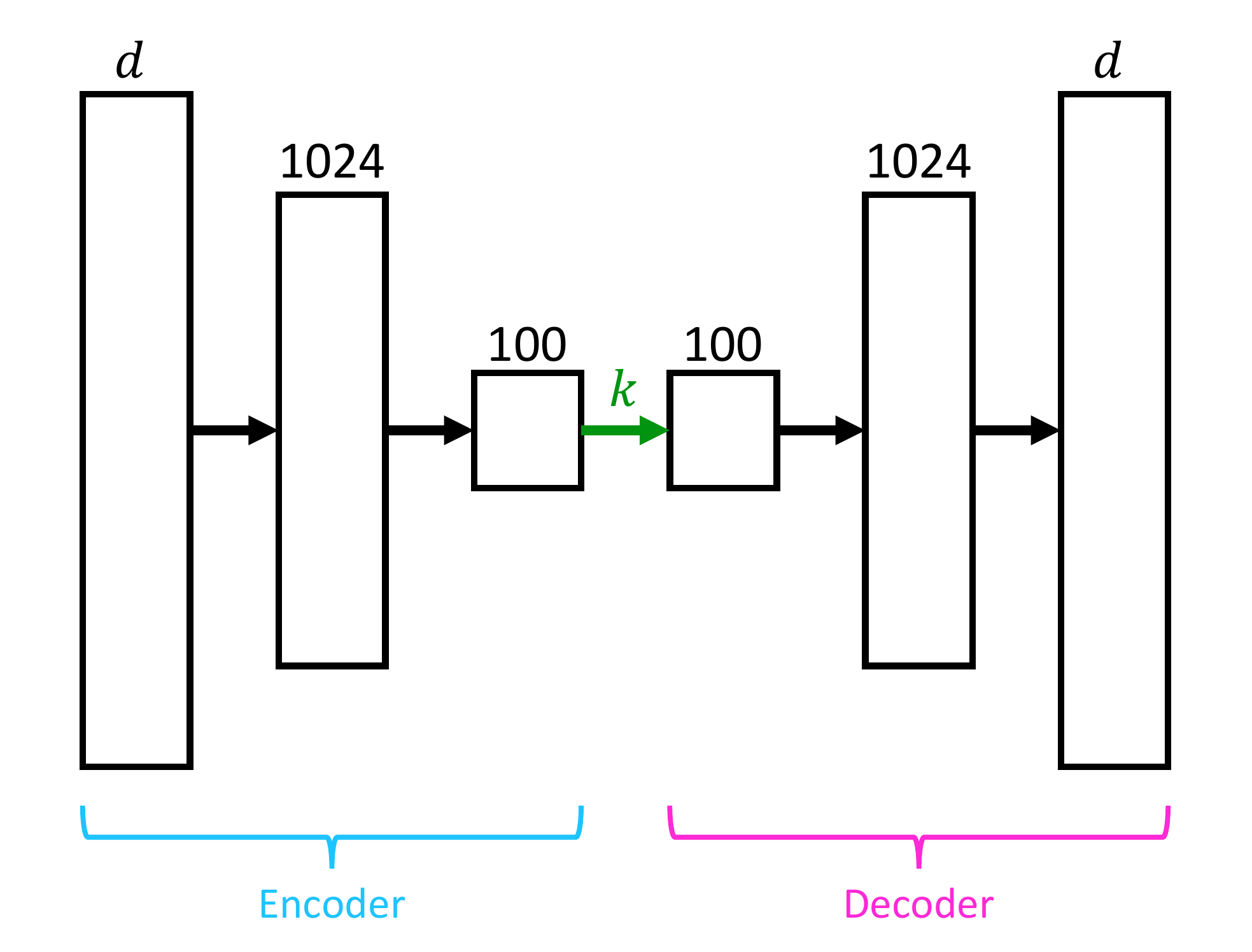}
  \caption{An outline of our autoencoders' design}
  \label{fig:autoencoder}
\end{figure}

An autoencoder (AE), as shown in \autoref{fig:autoencoder}, is another unsupervised learning method for reducing the dimensionality of input data and increasing the robustness of the representations learned by neural networks \cite{ng2011sparse}. It is succinctly comprised of two primary components, an encoder and a decoder. The goal is to find the learned code $\vec{z} \in \mathbb{R}^k$ that best compresses the inputs $\vec{x} \in \mathbb{R}^d$ where $k << d$. Notably, during the training phase, for the $i$-th sample $D_i$, the output is the same as the input (i.e., $D_i = (\vec{x}, \vec{x})$). Concerning our proceeding experiments, all clustering results obtained by using an autoencoder in an $\textit{a priori}$ manner can be found in Section \ref{sec:ae_cluster_vizs}.

\begin{figure}[t!]
    \centering
    \begin{subfigure}{0.48\textwidth}
    \includegraphics[width=\linewidth]{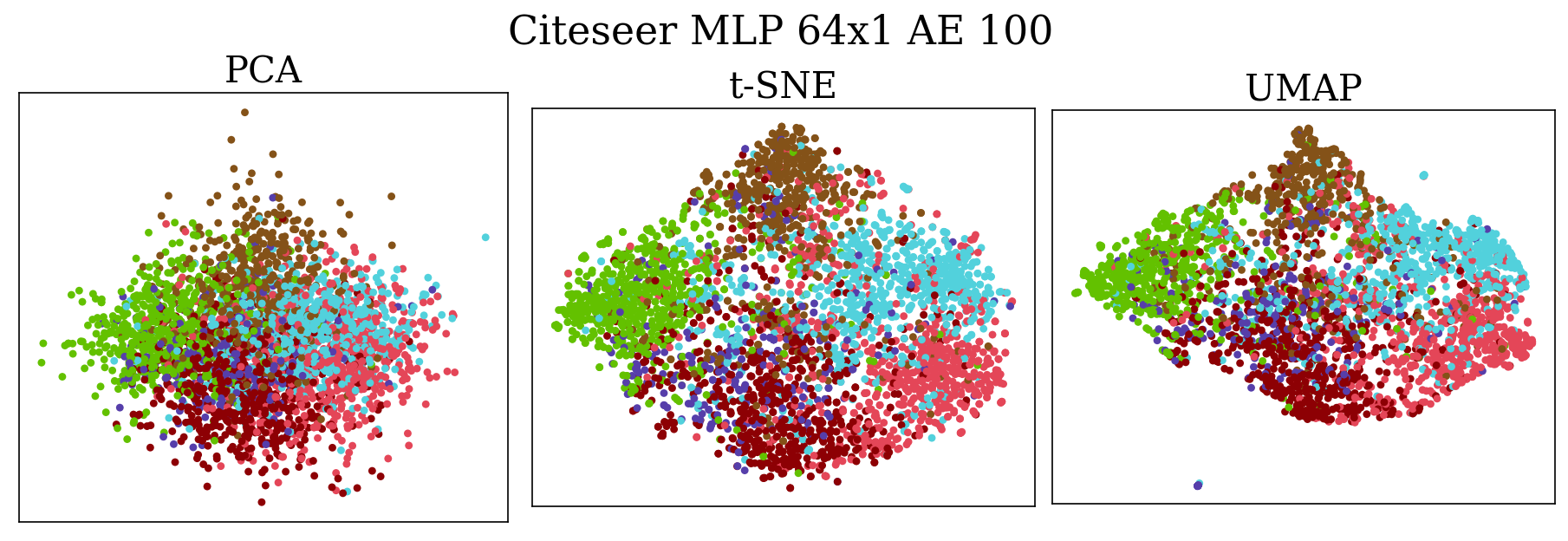}
    \includegraphics[width=\linewidth]{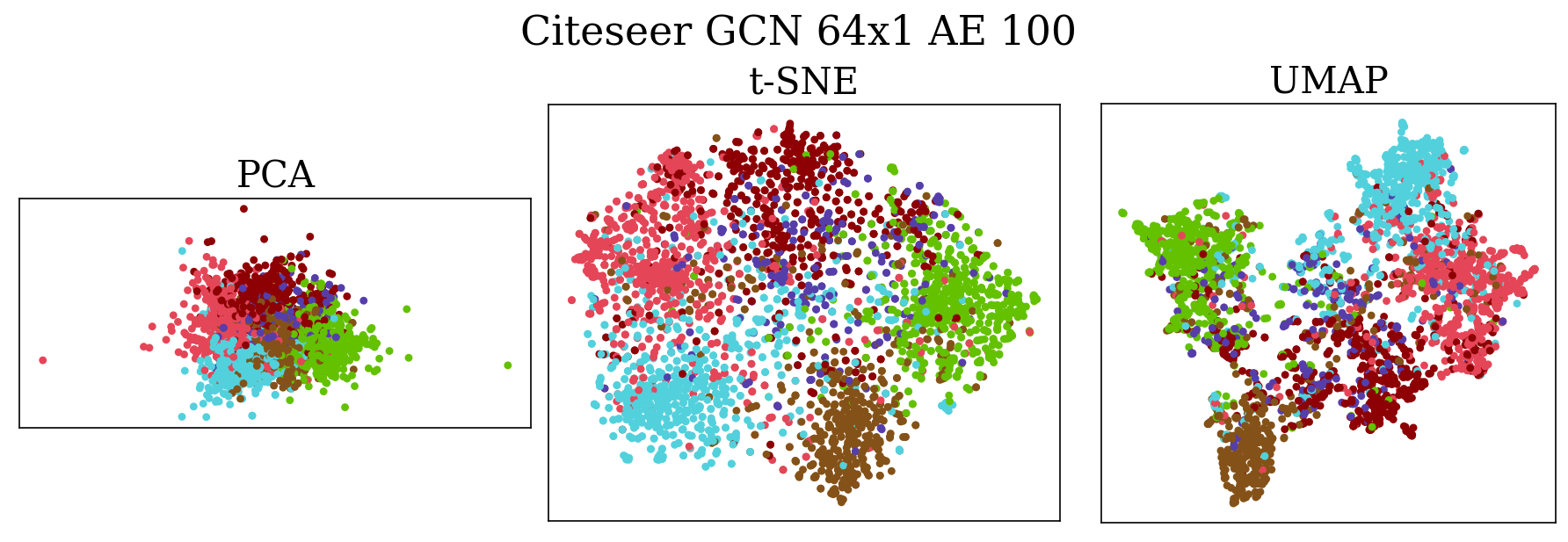}
    \caption{MLP and GCN clustering results with an AE applied $\textit{a priori}$.}
    \label{fig:citeseer_ap_ae_for_mlp_gcn}
    \end{subfigure}
    \begin{subfigure}{0.48\textwidth}
    \includegraphics[width=\linewidth]{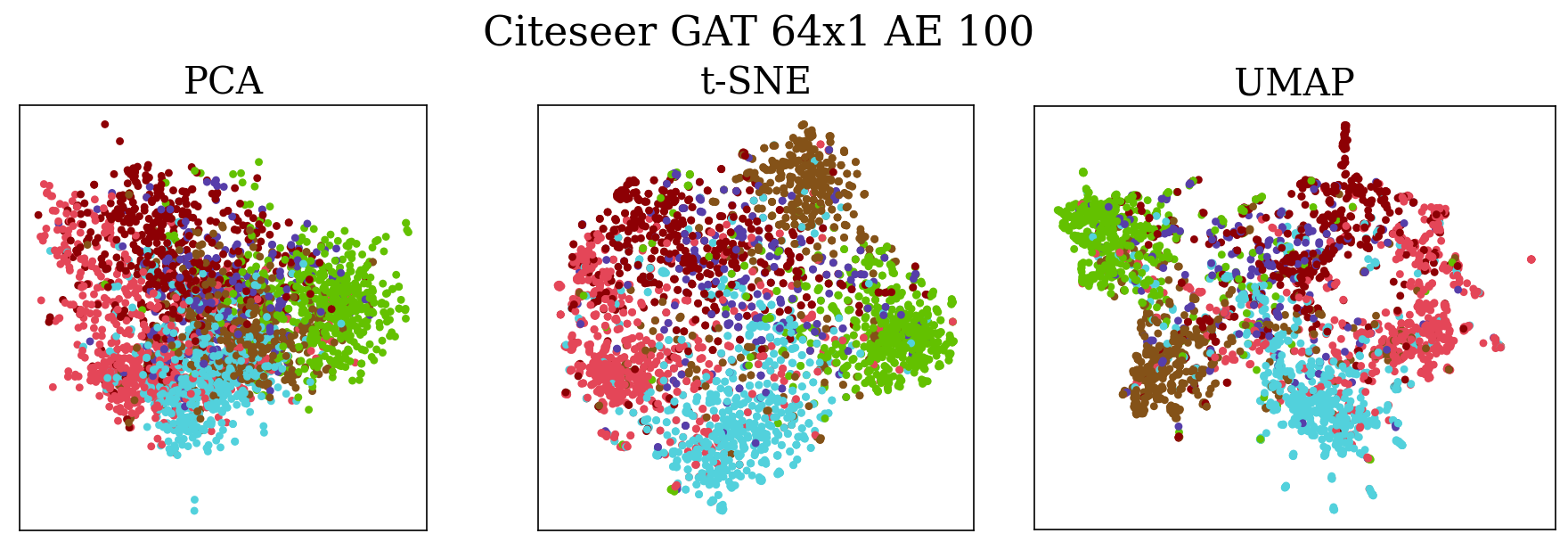}
    \includegraphics[width=\linewidth]{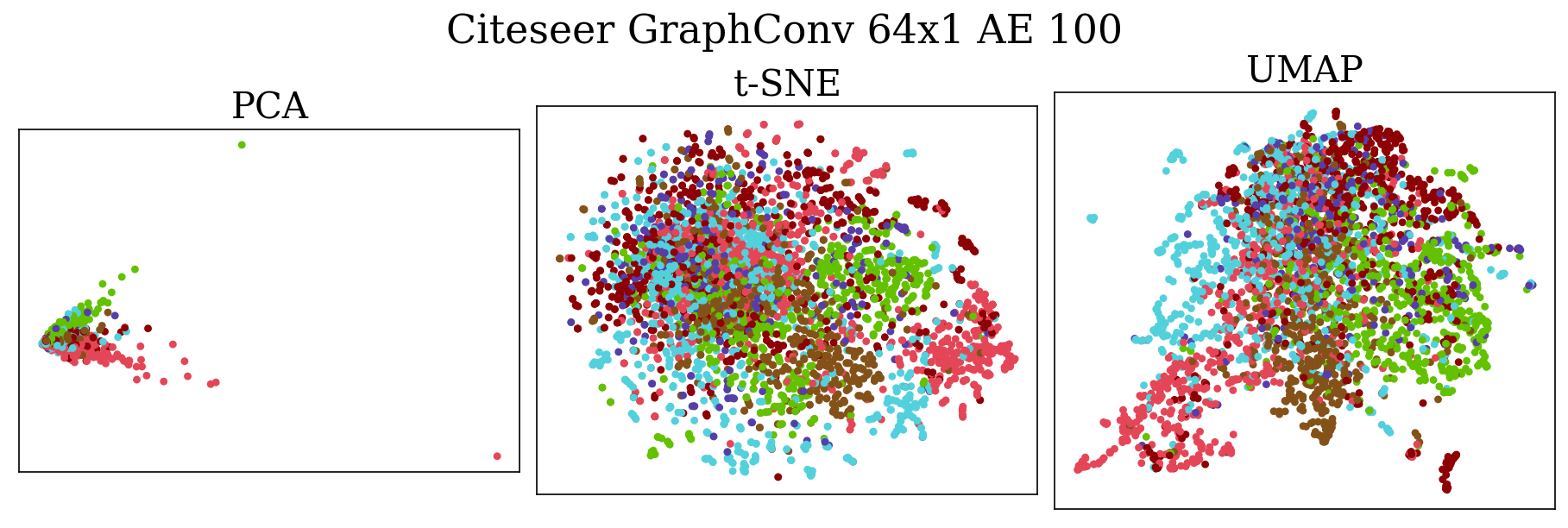}
    \caption{GAT and GraphConv clustering results with an AE applied $\textit{a priori}$.}
    \label{fig:citeseer_ap_ae_for_gat_graphconv}
    \end{subfigure}
    \caption{All models' clustering results on the Citeseer dataset with an AE applied $\textit{a priori}$.}
  \label{fig:ap_ae_mlp_gcn_gat_graphconv_on_citeseer}
\end{figure}

\section{Data}
In this work, we adopt the popular Cora \cite{10.5555/3041838.3041901} and Citeseer \cite{sen2008collective} node classification datasets for a series of reasons. First is their wide prior use in the GNN literature on SSL. We also chose these datasets to facilitate a transductive experimentation setting. For graph-based clustering and subsequent visualizations, it is often desirable to narrow focus to a single input graph for training, validation, and testing, which the setting of a transductive dataset directly provides.

The Cora dataset has 2,708 nodes (i.e., samples) and 5,429 edges in total. The number of training samples, validation samples, and test samples, are 140, 500, and 1,000, respectively. Since it has 1,433 unique words in its dictionary, one of its simplest forms of word representation is one-hot encoding. As such, the input features for the Cora dataset are in the set $\mathbb{R}^{1,433}$. The task of this dataset is to classify the subjects of scientific publications. There are seven subjects to be classified: (1) Case Based, (2) Genetic Algorithms, (3) Neural Networks, (4) Probabilistic Methods, (5) Reinforcement Learning, (6) Rule Learning, and (7) Theory.

\begin{figure}[b!]
    \centering
    \begin{subfigure}{0.48\textwidth}
    \includegraphics[width=\linewidth]{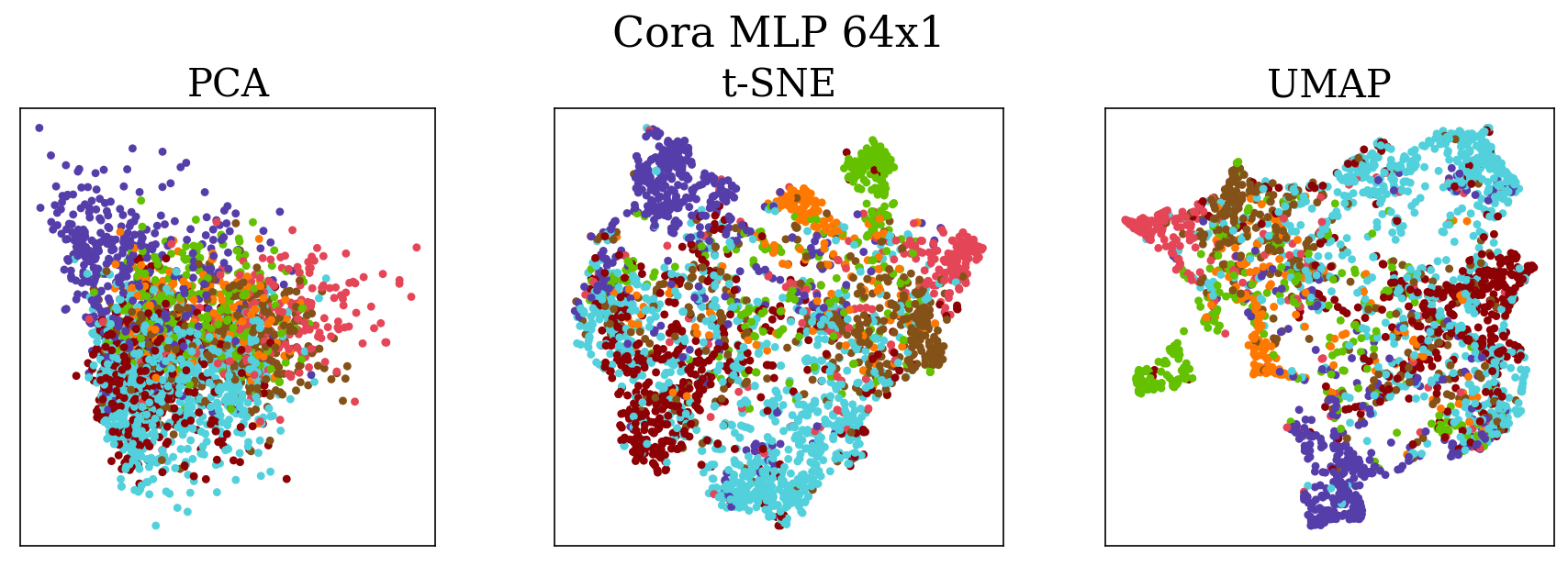}
    \includegraphics[width=\linewidth]{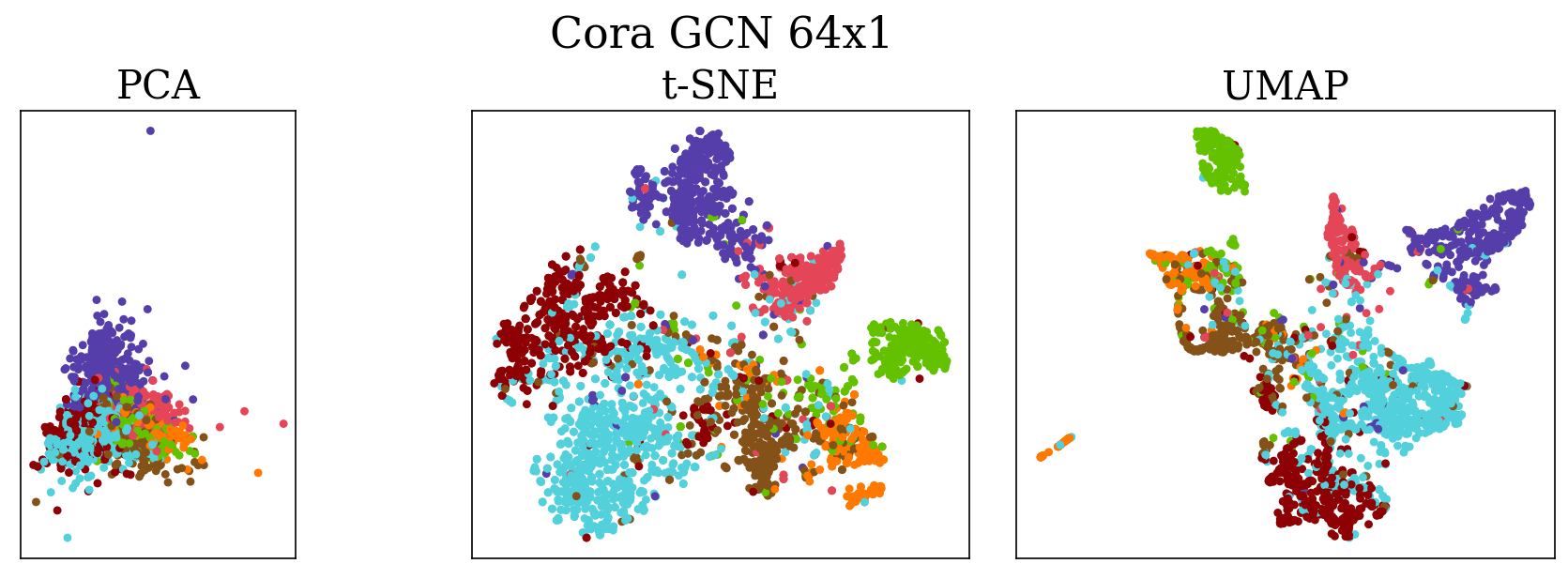}
    \caption{Without initial dimensionality reduction}
    \label{fig:cora_for_mlp_gcn}
    \end{subfigure}
    \begin{subfigure}{0.48\textwidth}
    \includegraphics[width=\linewidth]{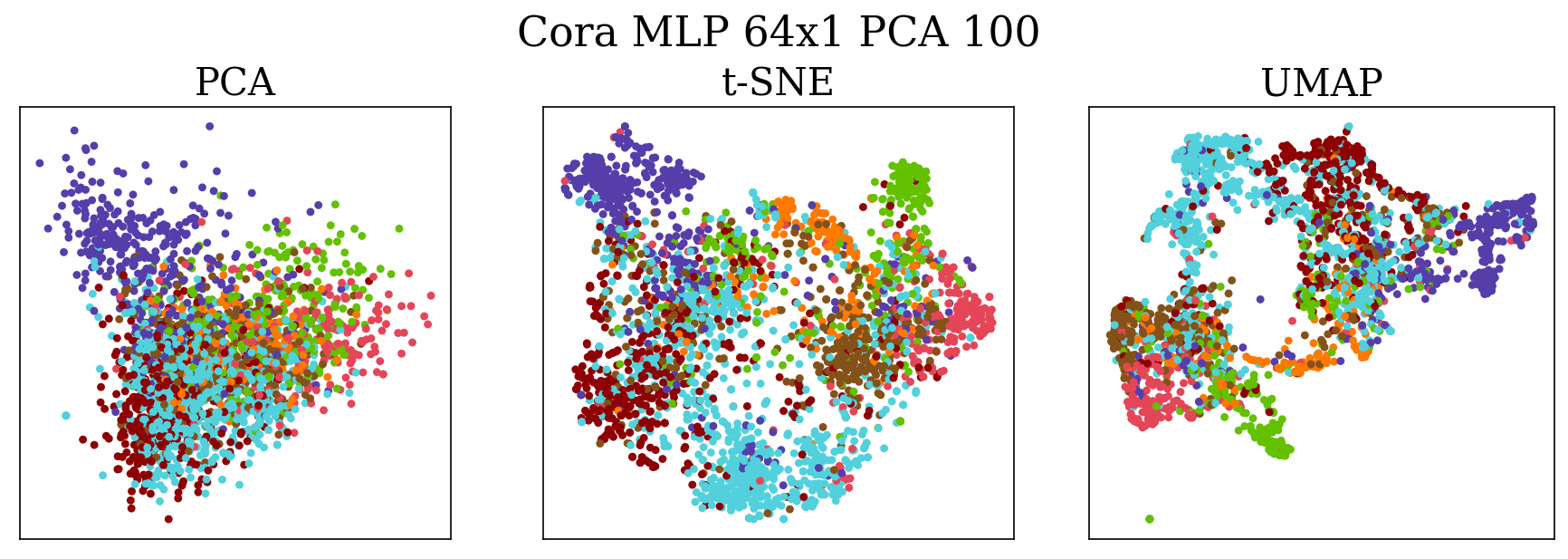}
    \includegraphics[width=\linewidth]{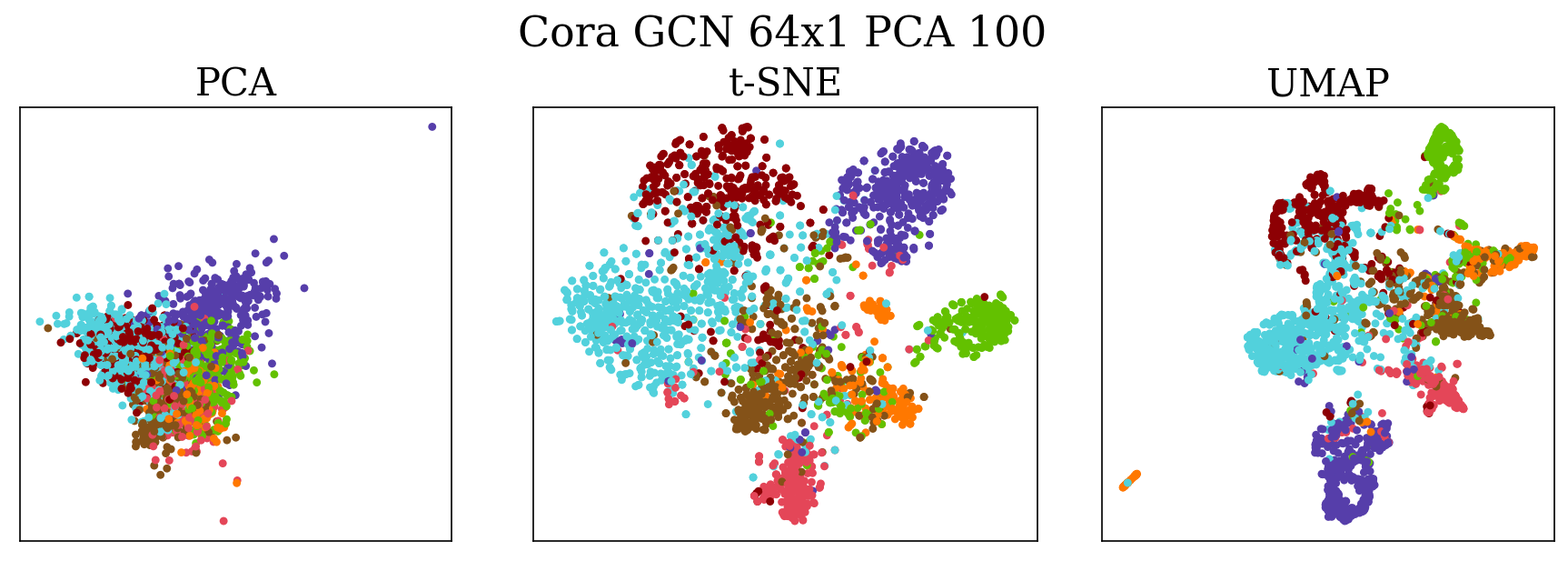}
    \caption{With PCA applied $\textit{a priori}$}
    \label{fig:cora_ap_pca_for_mlp_gcn}
    \end{subfigure}
    \caption{MLP and GCN clustering results on the Cora dataset.}
  \label{fig:mlp_gcn_on_cora}
\end{figure}

\begin{figure}[b!]
    \centering
    \begin{subfigure}{0.48\textwidth}
    \includegraphics[width=\linewidth]{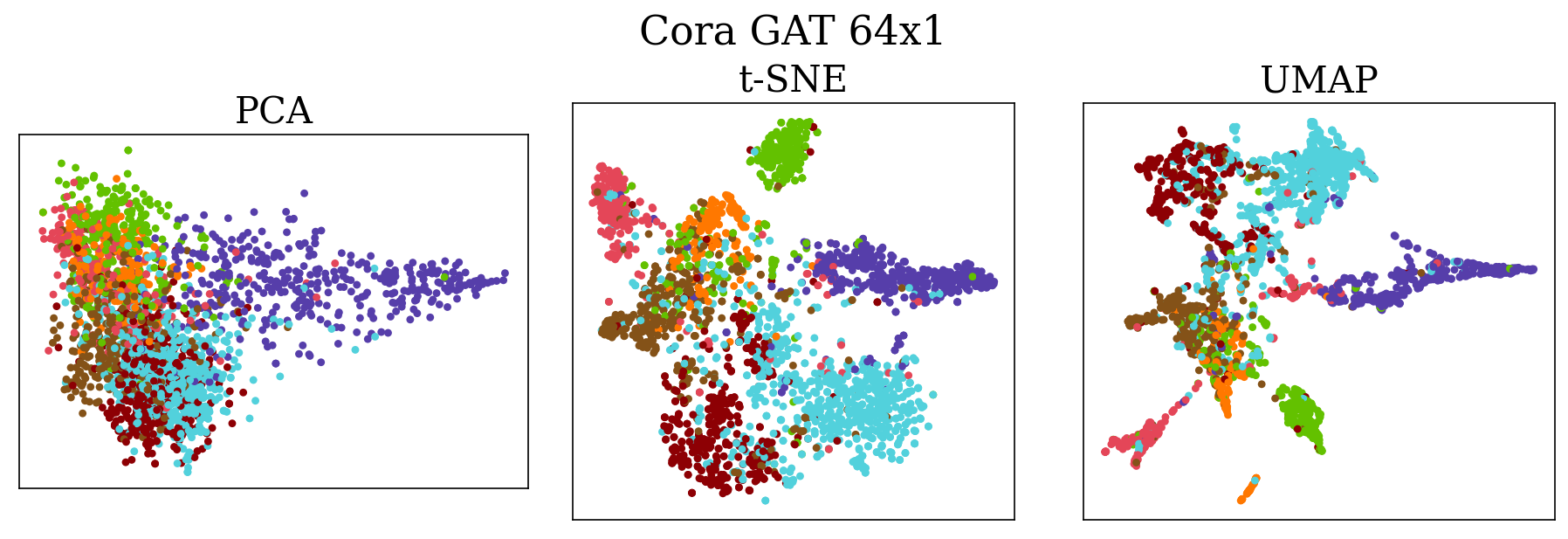}
    \includegraphics[width=\linewidth]{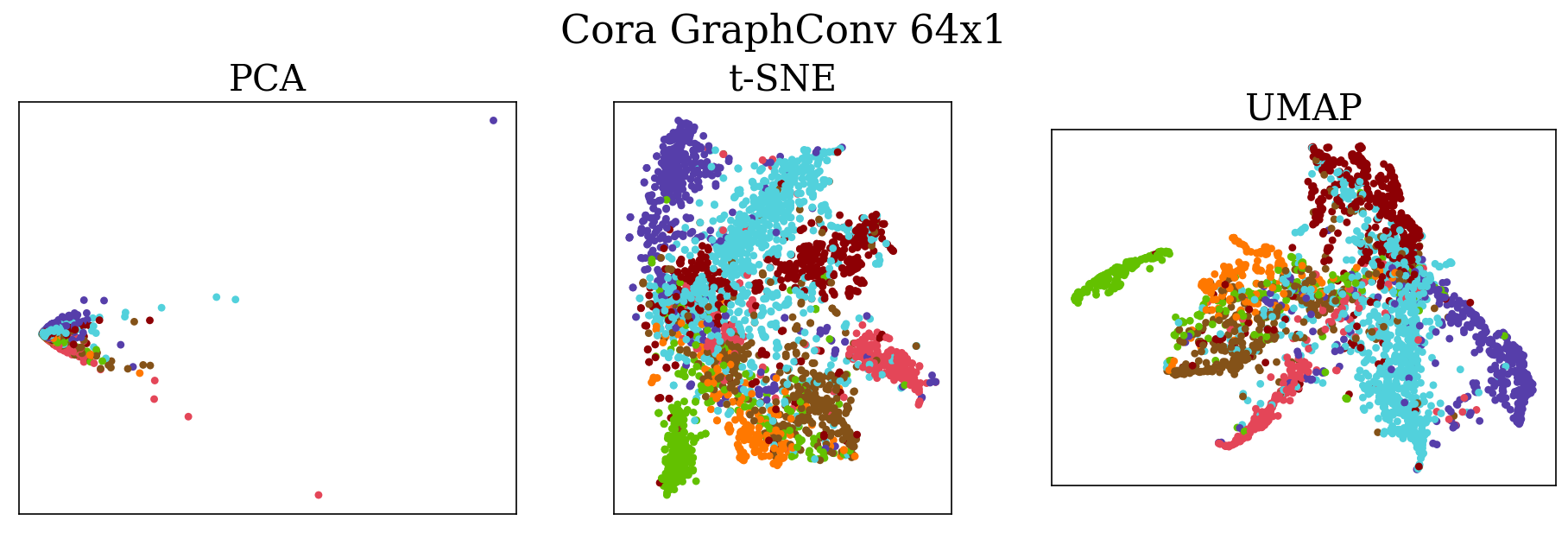}
    \caption{Without initial dimensionality reduction}
    \label{fig:cora_for_gat_graphconv}
    \end{subfigure}
    \begin{subfigure}{0.48\textwidth}
    \includegraphics[width=\linewidth]{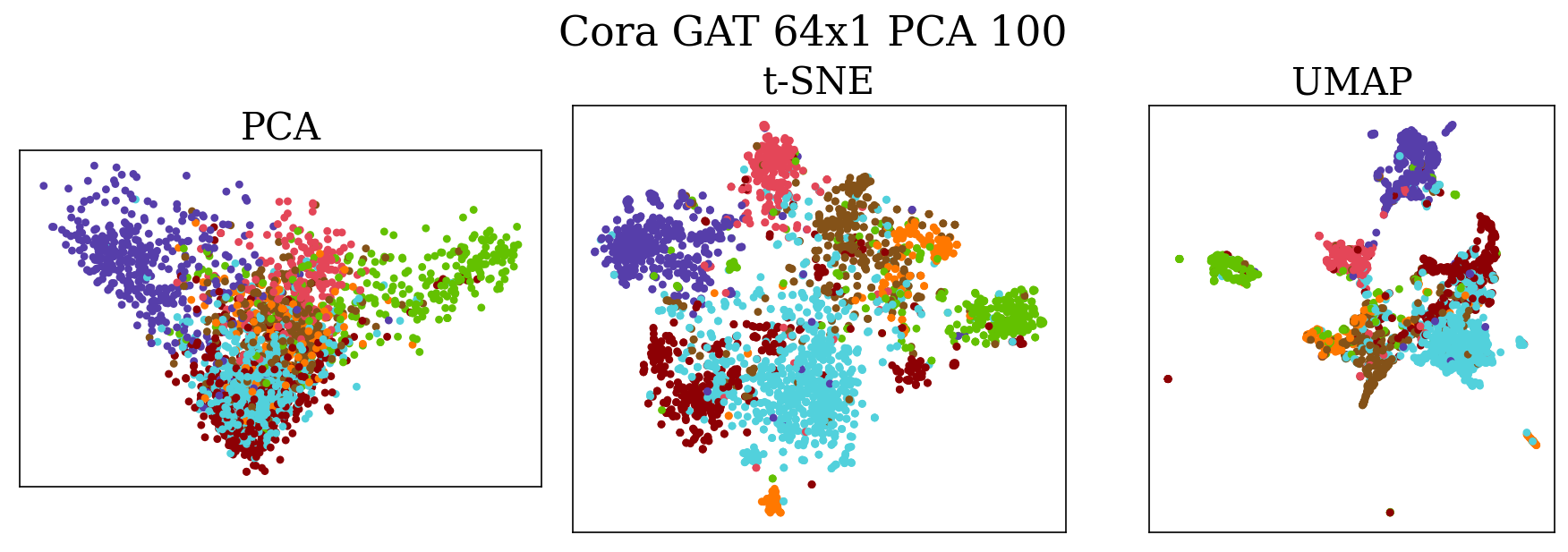}
    \includegraphics[width=\linewidth]{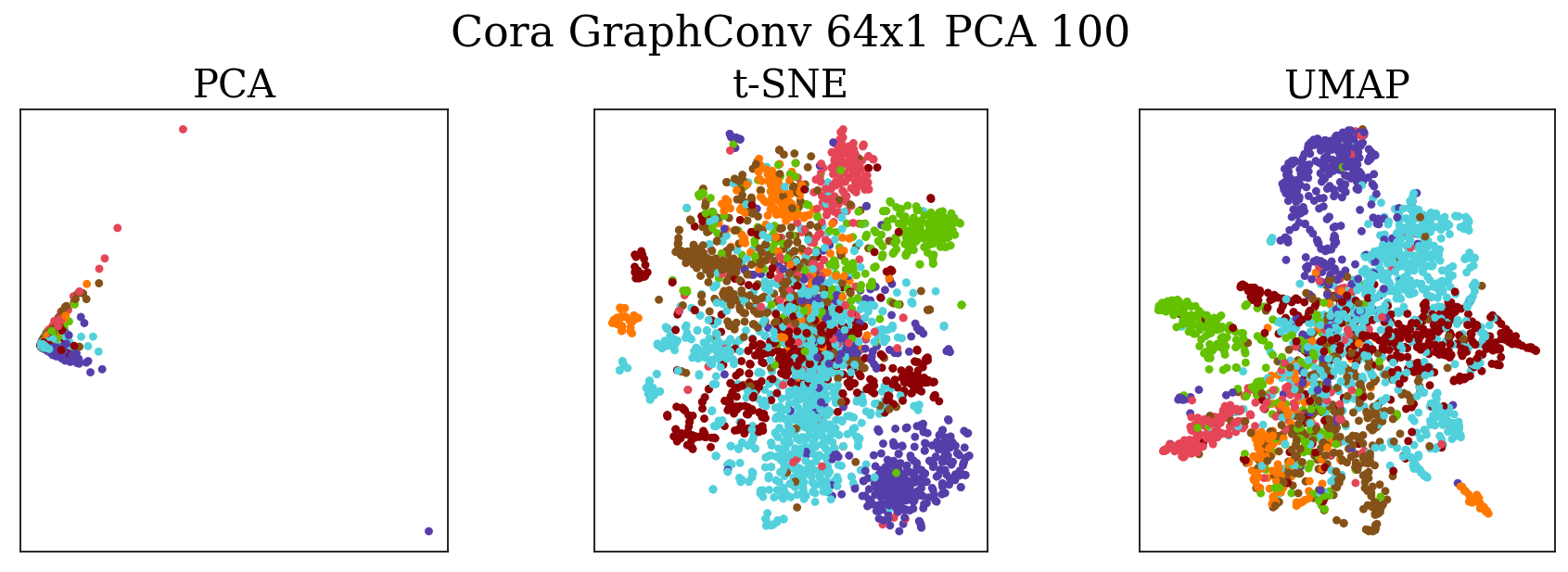}
    \caption{With PCA applied $\textit{a priori}$}
    \label{fig:cora_ap_pca_for_gat_graphconv}
    \end{subfigure}
    \caption{GAT and GraphConv clustering results on the Cora dataset.}
  \label{fig:gat_graphconv_on_cora}
\end{figure}

Similarly, the Citeseer dataset has 3,327 nodes (i.e., samples) and 9,228 edges in total. It is divided into three subsets, training set (120 samples), validation set (500 samples), and test set (1,000 samples). We utilize one-hot encoding with a size of 3,703 to represent the existence of words in the dataset's publications (i.e., nodes). This dataset also represents a classification problem, but there are now six classes consisting of publications related to (1) Agents, (2) Artificial Intelligence, (3) Database, (4) Information Retrieval, (5) Machine Learning, and (6) Human-Computer Interaction.

From the perspective of graph theory, each Cora publication is a node represented by its encoding in the space $\mathbb{R}^{1,433}$. Likewise, Citeseer uses the space $\mathbb{R}^{3,703}$ to represent its node encodings. This is suitable for GNNs that take graphs as inputs and produce new graphs out. For our semi-supervised study, node classifications on both datasets are performed to get subject predictions as our GNNs' final outputs for each node.

\section{Experiments}
\label{sec:experiments}

\subsection{Setup}

For all experiments conducted in our study, we used 1 hidden layer of the neural network chosen for the experiment and 64 intermediate channels to restrict the time required to train each model and to prevent the GNN models from oversmoothing their input node representations \cite{chen2020measuring}. We also arrived at the above configuration after experimenting with increased numbers of hidden layers (e.g., 2) for all models and observing $decreased$ classification performance and cluster quality. This means that a GNN with a single layer should be complex enough to solve this problem, for the purposes of our study. Subsequently, we used the stochastic gradient descent (SGD) optimizer \cite{bottou2010large} with the following, manually-tuned hyperparameters: a learning rate of $1e^{-1}$ (for all models except GraphConv, which required $1e^{-3}$ for training); a weight decay rate of $2e^{-3}$; SGD momentum of 0.9; a dropout (i.e., forget) rate \cite{srivastava2014dropout} of 0.1; and a batch size of 1. Additionally, with a multi-class cross entropy loss function for backpropagation \cite{zhang2018generalized}, we employed an early-stopping patience period of 5 epochs based on our validation loss \cite{yao2007early}.

\begin{figure}[t!]
    \centering
    \begin{subfigure}{0.48\textwidth}
    \includegraphics[width=\linewidth]{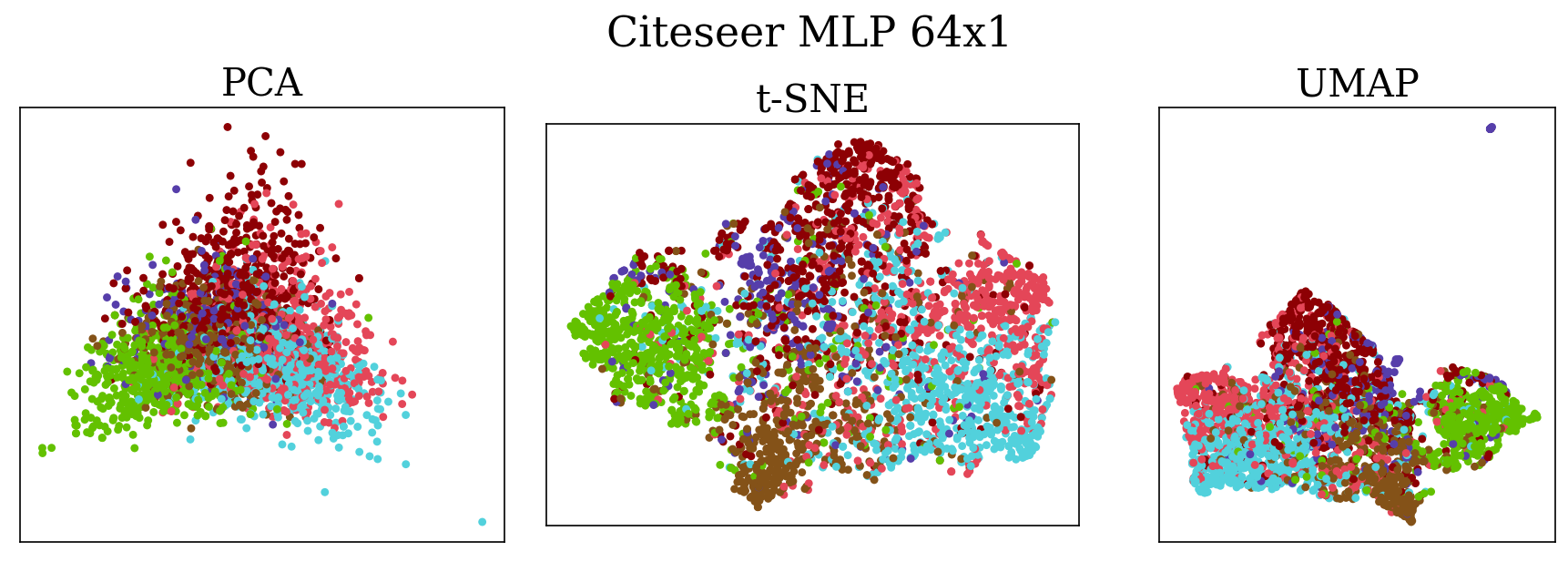}
    \includegraphics[width=\linewidth]{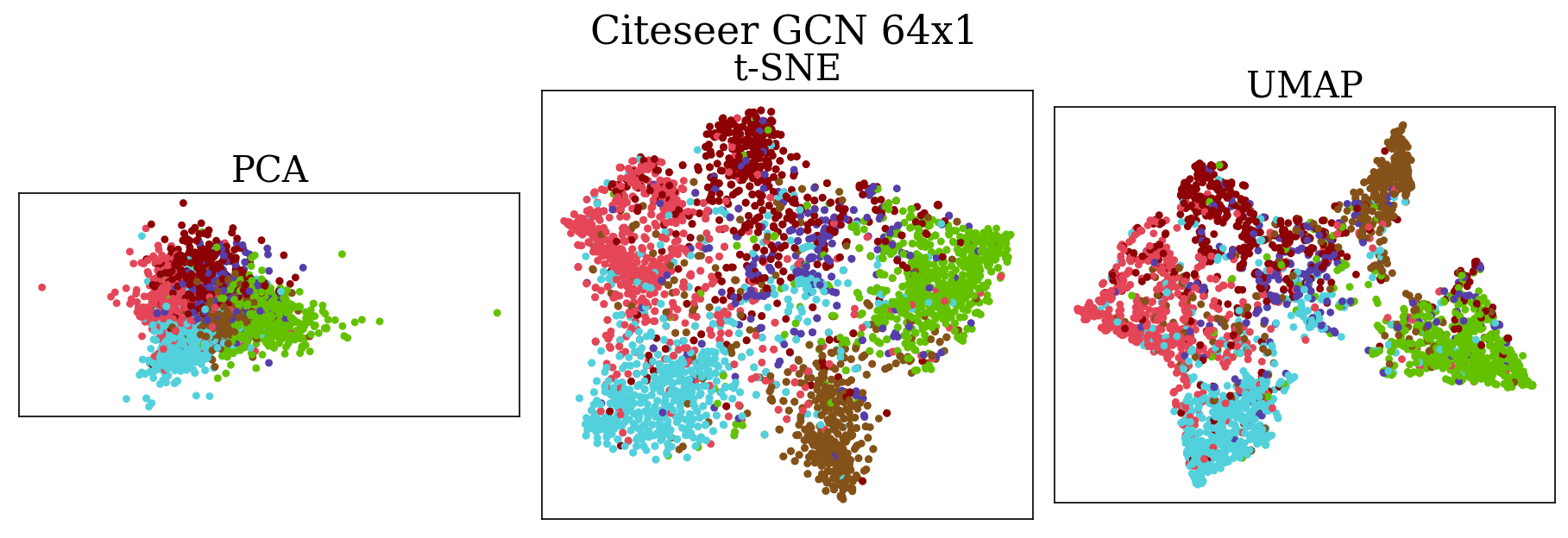}
    \caption{Without initial dimensionality reduction}
    \label{fig:citeseer_for_mlp_gcn}
    \end{subfigure}
    \begin{subfigure}{0.48\textwidth}
    \includegraphics[width=\linewidth]{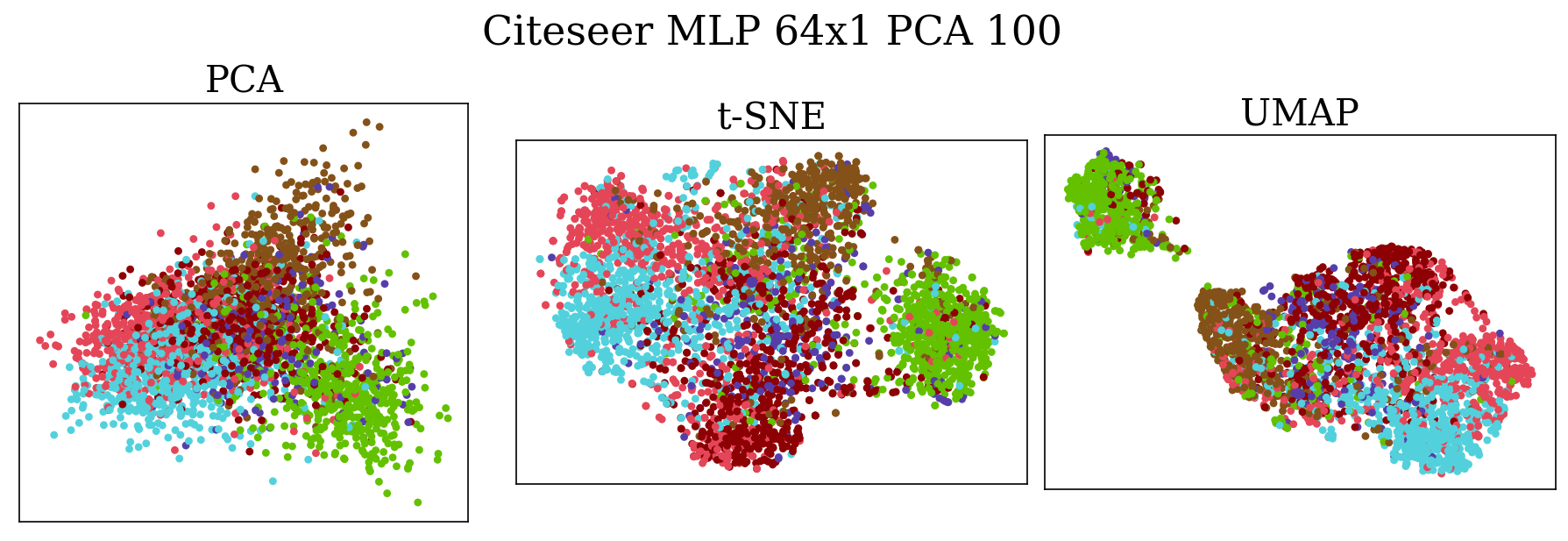}
    \includegraphics[width=\linewidth]{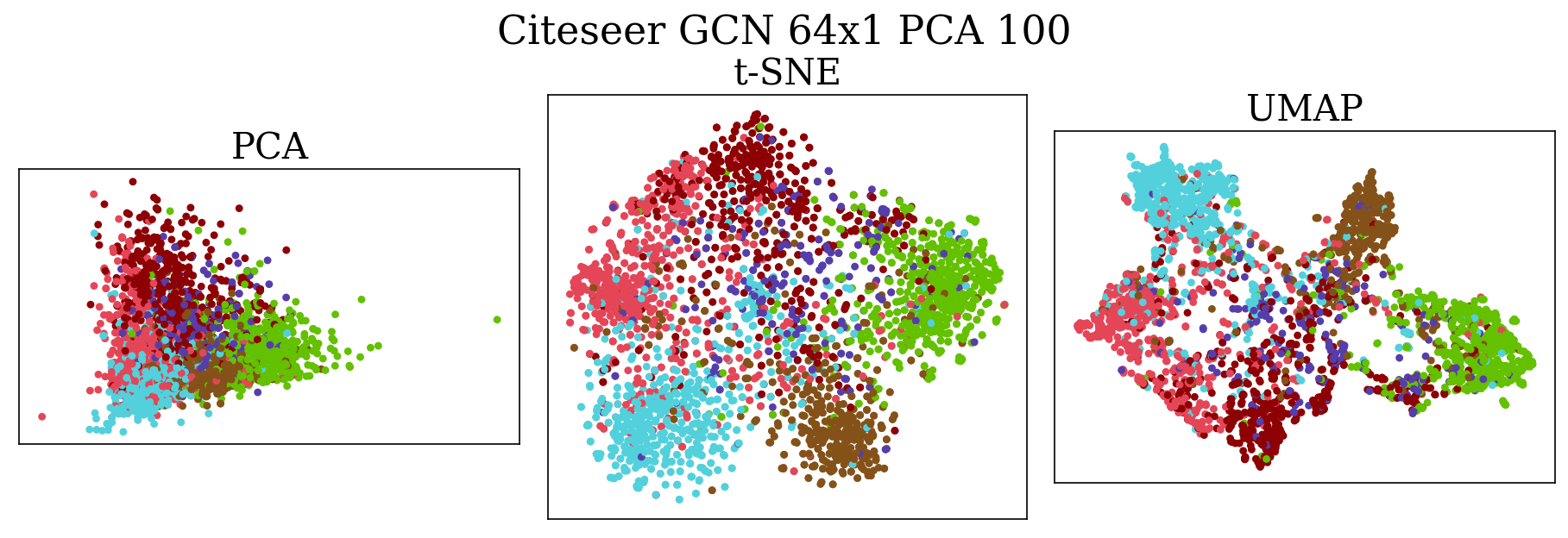}
    \caption{With PCA applied $\textit{a priori}$}
    \label{fig:citeseer_ap_pca_for_mlp_gcn}
    \end{subfigure}
    \caption{MLP and GCN clustering results on the Citeseer dataset.}
  \label{fig:mlp_gcn_on_citeseer}
\end{figure}

\begin{figure}[t!]
    \centering
    \begin{subfigure}{0.48\textwidth}
    \includegraphics[width=\linewidth]{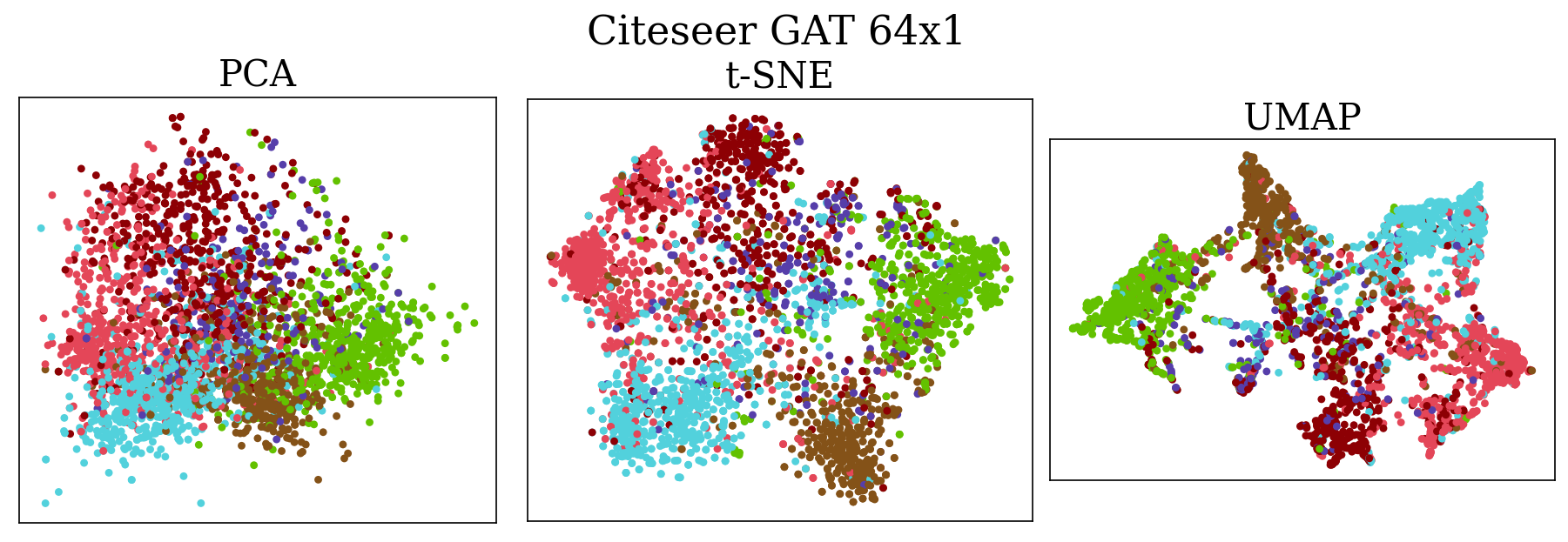}
    \includegraphics[width=\linewidth]{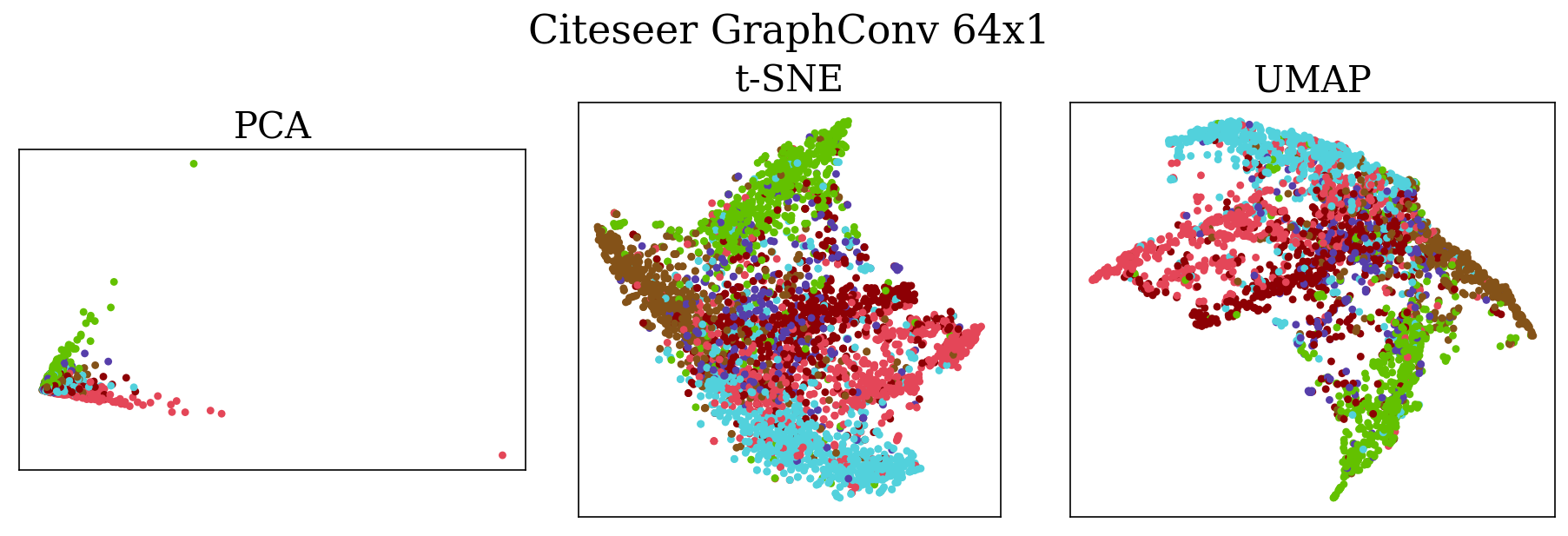}
    \caption{Without initial dimensionality reduction}
    \label{fig:citeseer_for_gat_graphconv}
    \end{subfigure}
    \begin{subfigure}{0.48\textwidth}
    \includegraphics[width=\linewidth]{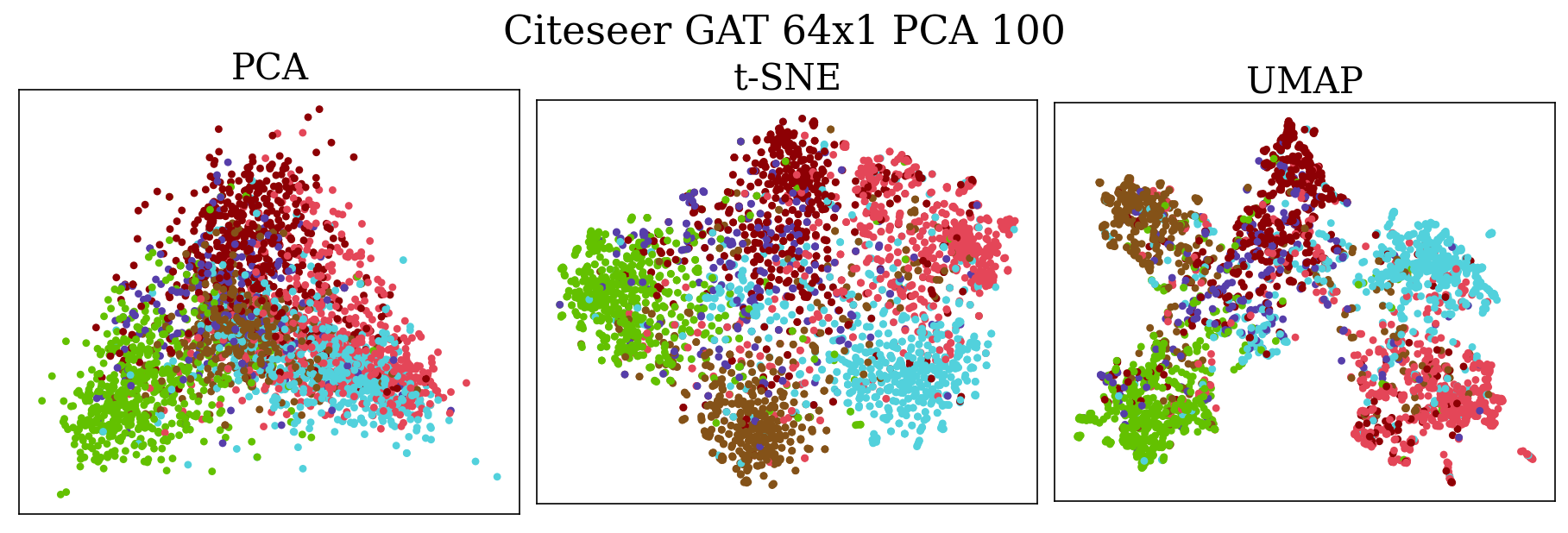}
    \includegraphics[width=\linewidth]{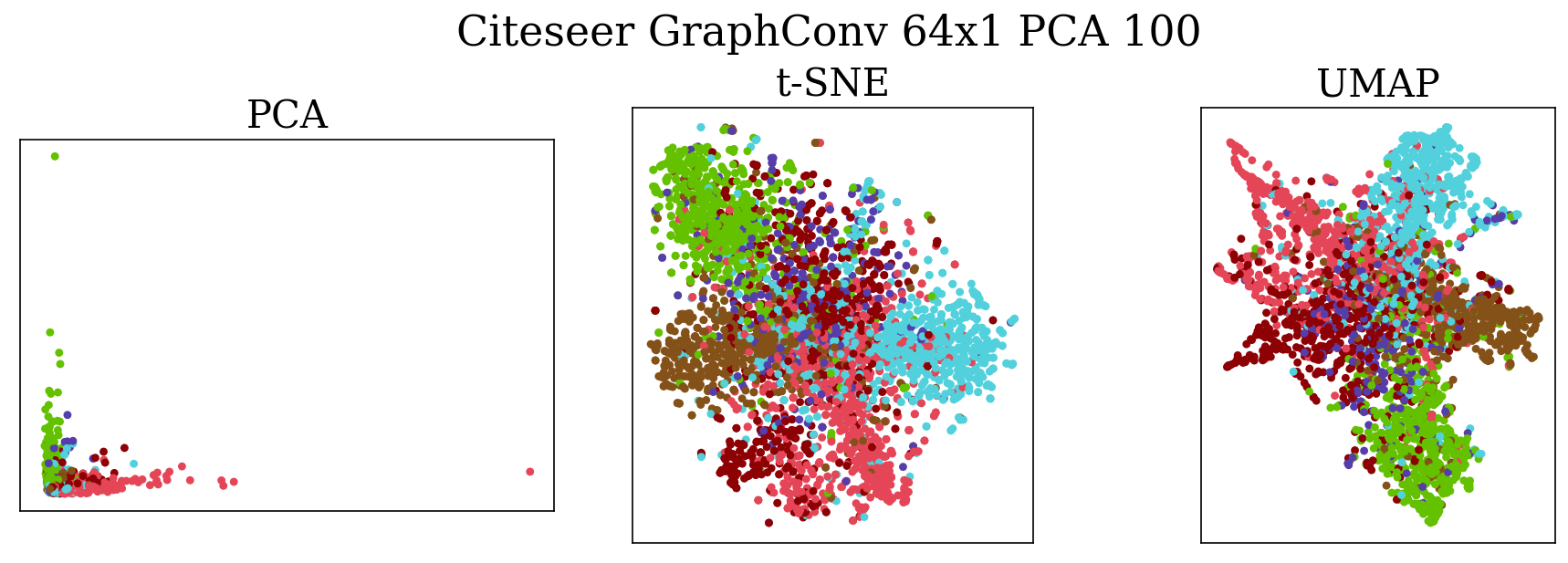}
    \caption{With PCA applied $\textit{a priori}$}
    \label{fig:citeseer_ap_pca_for_gat_graphconv}
    \end{subfigure}
    \caption{GAT and GraphConv clustering results on the Citeseer dataset.}
  \label{fig:gat_graphconv_on_citeseer}
\end{figure}

\subsection{Selection of Autoencoder Bottleneck Size}
In the context of training autoencoders, one needs to decide on the size of a bottleneck block to use. Consequently, we chose to validate our choice of bottleneck size by monitoring the mean squared error (MSE) we receive on our datasets' validation partitions with a fixed bottleneck size during the reconstruction of our datasets' original features. Once we identified the bottleneck size with the steepest decline in validation MSE (i.e., 100), which interestingly was the same for both of our datasets, that size became the default for all our experiments with autoencoders.

\subsection{Autoencoder Clustering Visualizations}
\label{sec:ae_cluster_vizs}

\begin{figure}[h]
    \centering
    \includegraphics[width=\linewidth]{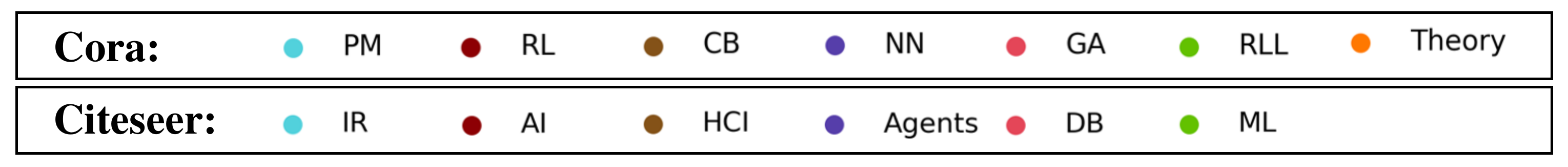}
    \caption{Data legend for the Cora dataset and the Citeseer dataset.}
    \label{fig:plots_legend}
\end{figure}

To explore the use of autoencoders for dimensionality reduction on graphs, we applied them in an $\textit{a priori}$ manner for each dataset during training. Using the node class-color legend illustrated in \autoref{fig:plots_legend}, based on our analysis Figures \ref{fig:ap_ae_mlp_gcn_gat_graphconv_on_cora} and \ref{fig:ap_ae_mlp_gcn_gat_graphconv_on_citeseer} demonstrate that $\textit{a priori}$ autoencoder-driven dimensionality reduction can produce reasonable and interpretable clustering visualizations for both datasets. As a consequence, in Sections \ref{sec:cora_results} and \ref{sec:citeseer_results}, we see that models trained on features that have had their dimensionality reduced by an autoencoder may also see enhanced clustering quality in quantitative terms. This suggests that exploring dimensionality reduction techniques adjacent to autoencoders, namely those specifically designed for graph-based data, may yield improved node clusterings.

\subsection{Cora Results}
\label{sec:cora_results}

\begin{table*}[t!]
    \caption{The effect of $\textit{a priori}$ dimensionality reduction on node classification for the Cora dataset. Results reported are means and standard deviations (in parentheses) of five separate runs with different random seeds.}
    \label{tbl:cora-class-benchmarks-table}
    \centering
        \begin{tabular}{lcccccr}
            \hline
            Model  &   Input   &   Accuracy   &   Precision   &   Recall    &   F1   \\
            \hline
            MLP & Original & 58.48 (0.73) & 56.08 (0.97) & 58.38 (0.64) & 56.53 (0.88) \\
            GCN & Original & 80.82 (0.62) & 79.47 (0.46) & 82.21 (0.60) & 80.39 (0.57) \\
            GAT & Original & 78.36 (0.47) & 76.58 (0.65) & 79.73 (0.30) & 77.68 (0.48) \\
            GraphConv & Original & 73.52 (1.52) & 72.43 (0.59) & 76.27 (1.27) & 73.34 (0.96) \\
            \hline
            MLP & PCA-100 & 57.90 (1.44) & 56.11 (1.76) & 58.75 (1.73) & 56.46 (1.77) \\
            GCN & PCA-100 & $\textbf{80.98}$ ($\textbf{0.80}$) & $\textbf{79.77}$ ($\textbf{0.92}$) & $\textbf{82.25}$ ($\textbf{0.70}$) & $\textbf{80.55}$ ($\textbf{0.86}$) \\
            GAT & PCA-100 & 78.36 (1.39) & 77.17 (1.46) & 79.78 (1.19) & 78.00 (1.31) \\
            GraphConv & PCA-100 & 72.04 (1.00) & 70.79 (0.57) & 75.09 (0.99) & 72.04 (0.70) \\
            \hline
            MLP & AE-100 & 63.24 (0.59) & 61.19 (0.61) & 63.21 (0.71) & 61.72 (0.55) \\
            GCN & AE-100 & 80.68 (0.45) & 78.84 (0.50) & 81.57 (0.51) & 79.80 (0.45) \\
            GAT & AE-100 & 77.64 (1.21) & 76.11 (0.95) & 79.16 (1.15) & 77.14 (1.05) \\
            GraphConv & AE-100 & 60.96 (11.21) & 60.01 (12.44) & 60.54 (14.63) & 58.72 (13.96) \\
            \hline
        \end{tabular}
\end{table*}

\begin{table*}[t!]
    \caption{The effect of $\textit{a priori}$ dimensionality reduction on node classification for the Citeseer dataset. Results reported are means and standard deviations (in parentheses) of five separate runs with different random seeds.}
    \label{tbl:citeseer-class-benchmarks-table}
    \centering
        \begin{tabular}{lcccccr}
            \hline
            Model  &   Input   &   Accuracy   &   Precision   &   Recall    &   F1   \\
            \hline
            MLP & Original & 56.56 (1.21) & 57.85 (0.89) & 56.05 (1.09) & 55.50 (1.10) \\
            GCN & Original & $\textbf{71.10}$ (0.84) & $\textbf{68.35}$ (0.82) & $\textbf{68.67}$ (0.89) & $\textbf{68.24}$ (0.86) \\
            GAT & Original & 68.84 (0.45) & 66.64 (0.43) & 66.68 (0.33) & 66.15 (0.36) \\
            GraphConv & Original & 65.64 (0.85) & 63.50 (0.43) & 62.83 (0.83) & 62.35 (0.75) \\
            \hline
            MLP & PCA-100 & 60.60 (0.94) & 59.51 (0.73) & 58.95 (0.94) & 58.59 (0.82) \\
            GCN & PCA-100 & $\textbf{70.24}$ ($\textbf{0.54}$) & $\textbf{67.65}$ ($\textbf{0.47}$) & $\textbf{67.71}$ ($\textbf{0.46}$) & $\textbf{67.12}$ ($\textbf{0.43}$) \\
            GAT & PCA-100 & 68.56 (0.84) & 65.87 (0.63) & 66.03 (0.68) & 65.56 (0.73) \\
            GraphConv & PCA-100 & 63.94 (1.00) & 61.41 (0.90) & 60.76 (1.15) & 60.35 (1.20) \\
            \hline
            MLP & AE-100 & 60.60 (1.39) & 60.30 (0.94) & 59.01 (1.11) & 58.61 (1.19) \\
            GCN & AE-100 & 69.14 (1.50) & 66.99 (0.97) & 66.83 (1.29) & 66.16 (1.37) \\
            GAT & AE-100 & 67.68 (0.77) & 66.19 (0.98) & 65.40 (0.90) & 64.87 (0.78) \\
            GraphConv & AE-100 & 45.68 (16.03) & 44.71 (18.36) & 45.06 (15.48) & 42.41 (17.87) \\
            \hline
        \end{tabular}
\end{table*}

Table \ref{tbl:cora-clust-benchmarks-table} portrays how for all our selected neural network models except that of \cite{morris2019weisfeiler}, applying either PCA or an autoencoder $\textit{a priori}$ favorably achieves low intra-cluster variance and high inter-cluster distance via both the Silhouette Coefficient \cite{zhu2010clustering} and the Dunn Index \cite{ncir2021parallel}. Particularly interesting to note is that, as Tables \ref{tbl:cora-class-benchmarks-table} and \ref{tbl:cora-clust-benchmarks-table} show, node clustering and node classification can be seen as complementary tasks. That is, in the context of the Cora dataset, node classification GNNs may improve their classification performance in conjunction with enhancements to their node clustering capability. In summary, we observe that the Cora dataset enables graph neural networks, particularly GCNs, to cluster and classify nodes in a concurrent and performant manner when applying PCA to node features in an $\textit{a priori}$ manner.

\subsection{Citeseer Results}
\label{sec:citeseer_results}

\begin{table}[htb]
    \caption{The effect of $\textit{a priori}$ dimensionality reduction on Cora dataset clusterings. Results reported are taken from the first of our five separate runs.}
    \label{tbl:cora-clust-benchmarks-table}
    \centering
        \begin{tabular}{lcccccr}
            \hline
            Model  &   Input    &   Output   &   Silhouette Coefficient   &   Dunn Index   \\
            \hline
            MLP & Original  &   PCA & -0.064 & 37.615 \\
            MLP & Original  &  t-SNE & -0.016 & 1.831 \\
            MLP & Original  &   UMAP & -0.019 & 18.202 \\
            GCN & Original  &   PCA & 0.012 & 18.496 \\
            GCN & Original  &  t-SNE & 0.212 & 1.388 \\
            GCN & Original  &   UMAP & 0.256 & 12.407 \\
            GAT & Original  &   PCA & -0.006 & 27.310 \\
            GAT & Original  &  t-SNE & 0.185 & 1.379 \\
            GAT & Original  &   UMAP & 0.200 & 12.127 \\
            GraphConv & Original  &   PCA & -0.097 & 3.097 \\
            GraphConv & Original  &  t-SNE & 0.047 & 1.571 \\
            GraphConv & Original  &   UMAP & 0.095 & 15.548 \\
            \hline
            MLP & PCA-100  &   PCA & -0.050 & $\mathbf{37.629}$ \\
            MLP & PCA-100  &  t-SNE & 0.004 & 1.844 \\
            MLP & PCA-100  &   UMAP & -0.018 & 14.851 \\
            GCN & PCA-100  &   PCA & -0.013 & 18.388 \\
            GCN & PCA-100  &  t-SNE & 0.195 & 1.365 \\
            GCN & PCA-100  &   UMAP & $\mathbf{0.227}$ & 11.888 \\
            GAT & PCA-100  &   PCA & 0.048 & 34.035 \\
            GAT & PCA-100  &  t-SNE & 0.154 & 1.242 \\
            GAT & PCA-100  &   UMAP & 0.208 & 11.649 \\
            GraphConv & PCA-100  &   PCA &  -0.111 & 3.584 \\
            GraphConv & PCA-100  &  t-SNE & 0.017 & 1.597 \\
            GraphConv & PCA-100  &   UMAP & 0.043 & 21.594 \\
            \hline
            MLP & AE-100  &   PCA & -0.047 & 29.394 \\
            MLP & AE-100  &  t-SNE & 0.039 & 1.588 \\
            MLP & AE-100  &   UMAP & 0.060 & 18.739 \\
            GCN & AE-100  &   PCA & 0.069 & 18.208 \\
            GCN & AE-100  &  t-SNE & 0.232 & 1.462 \\
            GCN & AE-100  &   UMAP & $\mathbf{0.269}$ & 12.372 \\
            GAT & AE-100  &   PCA & 0.032 & 25.572 \\
            GAT & AE-100  &  t-SNE & 0.175 & 1.259 \\
            GAT & AE-100  &   UMAP & 0.237 & 11.019 \\
            GraphConv & AE-100  &   PCA & -0.101 & 2.023 \\
            GraphConv & AE-100  &  t-SNE & -0.057 & 1.603 \\
            GraphConv & AE-100  &   UMAP & 0.062 & 23.875 \\
            \hline
        \end{tabular}
\end{table}

\begin{table}[htb]
    \caption{The effect of $\textit{a priori}$ dimensionality reduction on Citeseer dataset clusterings. Results reported are taken from the first of our five separate runs.}
    \label{tbl:citeseer-clust-benchmarks-table}
    \centering
        \begin{tabular}{lcccccr}
            \hline
            Model  &   Input    &   Output   &   Silhouette Coefficient   &   Dunn Index   \\
            \hline
            MLP & Original  &   PCA & -0.039 & 46.717 \\
            MLP & Original  &  t-SNE & 0.040 & 1.963 \\
            MLP & Original  &   UMAP & 0.053 & 16.762 \\
            GCN & Original  &   PCA & 0.063 & 27.347 \\
            GCN & Original  &  t-SNE & 0.155 & 1.661 \\
            GCN & Original  &   UMAP & 0.162 & 21.340 \\
            GAT & Original  &   PCA & 0.052 & 49.851 \\
            GAT & Original  &  t-SNE & 0.145 & 1.643 \\
            GAT & Original  &   UMAP & 0.169 & 20.164 \\
            GraphConv & Original  &   PCA & -0.149 & 4.420 \\
            GraphConv & Original  &  t-SNE & 0.071 & 2.018 \\
            GraphConv & Original  &   UMAP & 0.069 & 22.523 \\
            \hline
            MLP & PCA-100  &   PCA & -0.008 & $\mathbf{57.072}$ \\
            MLP & PCA-100  &  t-SNE & 0.054 & 2.120 \\
            MLP & PCA-100  &   UMAP & 0.067 & 21.253 \\
            GCN & PCA-100  &   PCA & 0.042 & 32.469 \\
            GCN & PCA-100  &  t-SNE & 0.154 & 1.636 \\
            GCN & PCA-100  &   UMAP & $\mathbf{0.174}$ & 23.572 \\
            GAT & PCA-100  &   PCA & 0.016 & 56.144 \\
            GAT & PCA-100  &  t-SNE & 0.141 & 1.619 \\
            GAT & PCA-100  &   UMAP & 0.144 & 21.416 \\
            GraphConv & PCA-100  &   PCA &  -0.150 & 5.830 \\
            GraphConv & PCA-100  &  t-SNE & 0.066 & 2.107 \\
            GraphConv & PCA-100  &   UMAP & 0.114 & 29.787 \\
            \hline
            MLP & AE-100  &   PCA & -0.003 & 34.902 \\
            MLP & AE-100  &  t-SNE & 0.072 & 1.906 \\
            MLP & AE-100  &   UMAP & 0.083 & 25.998 \\
            GCN & AE-100  &   PCA & 0.046 & 22.575 \\
            GCN & AE-100  &  t-SNE & 0.146 & 1.548 \\
            GCN & AE-100  &   UMAP & $\mathbf{0.155}$ & 24.821 \\
            GAT & AE-100  &   PCA & 0.031 & 37.711 \\
            GAT & AE-100  &  t-SNE & 0.143 & 1.707 \\
            GAT & AE-100  &   UMAP & 0.140 & 21.035 \\
            GraphConv & AE-100  &   PCA & -0.158 & 3.836 \\
            GraphConv & AE-100  &  t-SNE & -0.089 & 1.816 \\
            GraphConv & AE-100  &   UMAP & -0.047 & 30.334 \\
            \hline
        \end{tabular}
\end{table}

Parallel to our observations in Section \ref{sec:cora_results}, Table \ref{tbl:citeseer-clust-benchmarks-table} demonstrates that PCA applied $\textit{a priori}$ to the Citeseer dataset's node features results in high quality clusterings. Likewise, we observe in Table \ref{tbl:citeseer-class-benchmarks-table} that, although unaltered node features are sufficient for GNNs to achieve our highest observed classification metrics, $\textit{a priori}$ application of PCA enables them to more confidently label document nodes with their respective topics by reducing the variance of their classification performance across different random seeds used for training. As quantitative model confidence measures are desired components of machine learning research, we believe such results illustrate that dimensionality reduction can, in certain settings, reduce the stochasticity associated with the predictions made by GNNs.

\subsection{Model Parameter Efficiency when using Dimensionality Reduction}

\begin{table}[h]
    \caption{The number of trainable parameters of different combinations of models and inputs.}
    \label{tbl:num_params}
    \centering
        \begin{tabular}{lcrr}
            \hline
            \multirow{2}{*}{Model} & \multirow{2}{*}{Input} & \multicolumn{2}{c}{Number of Trainable Parameters}\\
            \cline{3-4}
            &  & Cora & Citeseer \\
            \hline
            MLP & Original & 23,063 & 59,366\\
            GCN & Original & 23,063 & 59,366\\
            GAT & Original & 23,109 & 59,410\\
            GraphConv & Original & 46,103 & 118,710\\
            \hline
            MLP & PCA-100/AE-100 & 1,735 & 1,718\\
            GCN & PCA-100/AE-100 & $\mathbf{1\mathrm{,}735}$ & $\mathbf{1\mathrm{,}718}$\\
            GAT & PCA-100/AE-100 & 1,781 & 1,762\\
            GraphConv & PCA-100/AE-100 & 3,447 & 3,414\\
            \hline
        \end{tabular}
\end{table}

A tertiary benefit of using dimensionality reduction during model training can come in terms of models' parameter efficiency. \autoref{tbl:num_params} displays the number of learnable parameters for each of our models according to each input feature size. We observe that after applying $\textit{a priori}$ dimensionality reduction techniques to lessen the dimensionality of our datasets' input features to 100, the number of learnable parameters required for training each of our models is greatly reduced. For the Cora dataset, all models trained with an $\textit{a priori}$ AE or PCA have approximately 13 times fewer trainable parameters compared to training them with Cora's original input feature size. Concerning the Citeseer dataset, applying dimensionality reduction to our models' input features decreases their learnable parameter counts by over a factor of 34. Both outcomes represent a noticeable improvement in terms of our models' size, computational complexity, and parameter efficiency obtained by applying $\textit{a priori}$ dimensionality reduction.

\subsection{Discussion}

We believe there are a few key insights we can distill from our study on dimensionality reduction in the setting of semi-supervised graph learning. Recall that, according to the curse of dimensionality, datasets with samples containing 1,433 and 3,703 dimensions should be considered very high dimensional datasets that are ripe for dimensionality reduction. As a case study for this phenomenon, our visualizations and quantitative results for the Cora dataset suggest that one may want to $\textit{a priori}$ apply either PCA or an autoencoder before using UMAP to visualize learned node features, and reduce the computation time required for training each model. In the context of the Citeseer dataset, we observe a similar trend in that $\textit{a priori}$ dimensionality reduction also positively impacts the node classification and node clustering performance of graph-based neural networks. Nonetheless, employing dimensionality reduction in the setting of graph-based deep learning (particularly semi-supervised graph learning) requires one to consider the underlying characteristics of the dataset being analyzed such as its size and data diversity, as such characteristics may determine whether the use of $\textit{a priori}$ dimensionality reduction is appropriate and desirable.

\section{Conclusion}

In this work, we have empirically and visually shown that a careful application of $\textit{a priori}$ dimensionality reduction, namely autoencoding and PCA, can improve cluster quality and label propagation for enhanced node classification on graph datasets such as Cora while exploring other benefits of dimensionality reduction in GNN-based node clustering. For future work, we believe performing experiments with $\textit{a priori}$ t-SNE and UMAP applied to initial node features may promote further discussions into the characteristics of each dimensionality reduction algorithm for clustering and classification in a graph-transductive setting. Furthermore, we believe expanding the breadth of our study's transductive graph datasets is likely to yield more insights into how each dataset is best structured for improved pattern recognition. Investigations into alternatives to the autoencoder architecture we used in our study, such as graph variational autoencoders or graph normalizing flows, may provide the groundwork for new studies into advanced GNN-based clustering methods. Finally, we believe that exploring the use of newer graph neural networks for semi-supervised graph learning, networks such as the UniMP and GCNII models, may enrich the contents of future studies and the insights they provide.

\bibliography{references}

\begin{thebibliography}{10}
\providecommand{\url}[1]{#1}
\csname url@samestyle\endcsname
\providecommand{\newblock}{\relax}
\providecommand{\bibinfo}[2]{#2}
\providecommand{\BIBentrySTDinterwordspacing}{\spaceskip=0pt\relax}
\providecommand{\BIBentryALTinterwordstretchfactor}{4}
\providecommand{\BIBentryALTinterwordspacing}{\spaceskip=\fontdimen2\font plus
\BIBentryALTinterwordstretchfactor\fontdimen3\font minus
  \fontdimen4\font\relax}
\providecommand{\BIBforeignlanguage}[2]{{%
\expandafter\ifx\csname l@#1\endcsname\relax
\typeout{** WARNING: IEEEtran.bst: No hyphenation pattern has been}%
\typeout{** loaded for the language `#1'. Using the pattern for}%
\typeout{** the default language instead.}%
\else
\language=\csname l@#1\endcsname
\fi
#2}}
\providecommand{\BIBdecl}{\relax}
\BIBdecl

\bibitem{van2020survey}
J.~E. Van~Engelen and H.~H. Hoos, ``A survey on semi-supervised learning,''
  \emph{Machine Learning}, vol. 109, no.~2, pp. 373--440, 2020.

\bibitem{xu2020label}
B.~Xu, J.~Huang, L.~Hou, H.~Shen, J.~Gao, and X.~Cheng, ``Label-consistency
  based graph neural networks for semi-supervised node classification,'' in
  \emph{Proceedings of the 43rd International ACM SIGIR conference on research
  and development in Information Retrieval}, 2020, pp. 1897--1900.

\bibitem{kang2021structured}
Z.~Kang, C.~Peng, Q.~Cheng, X.~Liu, X.~Peng, Z.~Xu, and L.~Tian, ``Structured
  graph learning for clustering and semi-supervised classification,''
  \emph{Pattern Recognition}, vol. 110, p. 107627, 2021.

\bibitem{reddy2020analysis}
G.~T. Reddy, M.~P.~K. Reddy, K.~Lakshmanna, R.~Kaluri, D.~S. Rajput,
  G.~Srivastava, and T.~Baker, ``Analysis of dimensionality reduction
  techniques on big data,'' \emph{IEEE Access}, vol.~8, pp. 54\,776--54\,788,
  2020.

\bibitem{10.5555/3041838.3041901}
Q.~Lu and L.~Getoor, ``Link-based classification,'' in \emph{Proceedings of the
  Twentieth International Conference on International Conference on Machine
  Learning}, ser. ICML'03.\hskip 1em plus 0.5em minus 0.4em\relax AAAI Press,
  2003, p. 496–503.

\bibitem{sen2008collective}
P.~Sen, G.~Namata, M.~Bilgic, L.~Getoor, B.~Galligher, and T.~Eliassi-Rad,
  ``Collective classification in network data,'' \emph{AI magazine}, vol.~29,
  no.~3, pp. 93--93, 2008.

\bibitem{nigam2000text}
K.~Nigam, A.~K. McCallum, S.~Thrun, and T.~Mitchell, ``Text classification from
  labeled and unlabeled documents using em,'' \emph{Machine learning}, vol.~39,
  no.~2, pp. 103--134, 2000.

\bibitem{chapelle2006continuation}
O.~Chapelle, M.~Chi, and A.~Zien, ``A continuation method for semi-supervised
  svms,'' in \emph{Proceedings of the 23rd international conference on Machine
  learning}, 2006, pp. 185--192.

\bibitem{berthelot2019mixmatch}
D.~Berthelot, N.~Carlini, I.~Goodfellow, N.~Papernot, A.~Oliver, and C.~Raffel,
  ``Mixmatch: A holistic approach to semi-supervised learning,'' \emph{arXiv
  preprint arXiv:1905.02249}, 2019.

\bibitem{DBLP:journals/corr/KipfW16}
\BIBentryALTinterwordspacing
T.~N. Kipf and M.~Welling, ``Semi-supervised classification with graph
  convolutional networks,'' \emph{CoRR}, vol. abs/1609.02907, 2016. [Online].
  Available: \url{http://arxiv.org/abs/1609.02907}
\BIBentrySTDinterwordspacing

\bibitem{yang2016revisiting}
Z.~Yang, W.~Cohen, and R.~Salakhudinov, ``Revisiting semi-supervised learning
  with graph embeddings,'' in \emph{International conference on machine
  learning}.\hskip 1em plus 0.5em minus 0.4em\relax PMLR, 2016, pp. 40--48.

\bibitem{van2009dimensionality}
L.~Van Der~Maaten, E.~Postma, J.~Van~den Herik \emph{et~al.}, ``Dimensionality
  reduction: a comparative,'' \emph{J Mach Learn Res}, vol.~10, no. 66-71,
  p.~13, 2009.

\bibitem{yan2006graph}
S.~Yan, D.~Xu, B.~Zhang, H.-J. Zhang, Q.~Yang, and S.~Lin, ``Graph embedding
  and extensions: A general framework for dimensionality reduction,''
  \emph{IEEE transactions on pattern analysis and machine intelligence},
  vol.~29, no.~1, pp. 40--51, 2006.

\bibitem{zhang2012graph}
L.~Zhang, S.~Chen, and L.~Qiao, ``Graph optimization for dimensionality
  reduction with sparsity constraints,'' \emph{Pattern Recognition}, vol.~45,
  no.~3, pp. 1205--1210, 2012.

\bibitem{mao2015dimensionality}
Q.~Mao, L.~Wang, S.~Goodison, and Y.~Sun, ``Dimensionality reduction via graph
  structure learning,'' in \emph{Proceedings of the 21th ACM SIGKDD
  international conference on knowledge discovery and data mining}, 2015, pp.
  765--774.

\bibitem{zhao2020connecting}
L.~Zhao and L.~Akoglu, ``Connecting graph convolutional networks and
  graph-regularized pca,'' \emph{arXiv preprint arXiv:2006.12294}, 2020.

\bibitem{goodfellow2016deep}
I.~Goodfellow, Y.~Bengio, and A.~Courville, \emph{Deep learning}.\hskip 1em
  plus 0.5em minus 0.4em\relax MIT press, 2016.

\bibitem{velivckovic2017graph}
P.~Veli{\v{c}}kovi{\'c}, G.~Cucurull, A.~Casanova, A.~Romero, P.~Lio, and
  Y.~Bengio, ``Graph attention networks,'' \emph{arXiv preprint
  arXiv:1710.10903}, 2017.

\bibitem{morris2019weisfeiler}
C.~Morris, M.~Ritzert, M.~Fey, W.~L. Hamilton, J.~E. Lenssen, G.~Rattan, and
  M.~Grohe, ``Weisfeiler and leman go neural: Higher-order graph neural
  networks,'' in \emph{Proceedings of the AAAI Conference on Artificial
  Intelligence}, vol.~33, no.~01, 2019, pp. 4602--4609.

\bibitem{chen2007enhanced}
X.-w. Chen and J.~C. Jeong, ``Enhanced recursive feature elimination,'' in
  \emph{Sixth International Conference on Machine Learning and Applications
  (ICMLA 2007)}.\hskip 1em plus 0.5em minus 0.4em\relax IEEE, 2007, pp.
  429--435.

\bibitem{doi:10.1080/14786440109462720}
\BIBentryALTinterwordspacing
K.~P. F.R.S., ``Liii. on lines and planes of closest fit to systems of points
  in space,'' \emph{The London, Edinburgh, and Dublin Philosophical Magazine
  and Journal of Science}, vol.~2, no.~11, pp. 559--572, 1901. [Online].
  Available: \url{https://doi.org/10.1080/14786440109462720}
\BIBentrySTDinterwordspacing

\bibitem{JMLR:v9:vandermaaten08a}
\BIBentryALTinterwordspacing
L.~van~der Maaten and G.~Hinton, ``Visualizing data using t-sne,''
  \emph{Journal of Machine Learning Research}, vol.~9, no.~86, pp. 2579--2605,
  2008. [Online]. Available:
  \url{http://jmlr.org/papers/v9/vandermaaten08a.html}
\BIBentrySTDinterwordspacing

\bibitem{van2008visualizing}
L.~Van~der Maaten and G.~Hinton, ``Visualizing data using t-sne.''
  \emph{Journal of machine learning research}, vol.~9, no.~11, 2008.

\bibitem{van2014renyi}
T.~Van~Erven and P.~Harremos, ``R{\'e}nyi divergence and kullback-leibler
  divergence,'' \emph{IEEE Transactions on Information Theory}, vol.~60, no.~7,
  pp. 3797--3820, 2014.

\bibitem{mcinnes2018umap}
L.~McInnes, J.~Healy, and J.~Melville, ``Umap: Uniform manifold approximation
  and projection for dimension reduction,'' \emph{arXiv preprint
  arXiv:1802.03426}, 2018.

\bibitem{becht2019dimensionality}
E.~Becht, L.~McInnes, J.~Healy, C.-A. Dutertre, I.~W. Kwok, L.~G. Ng,
  F.~Ginhoux, and E.~W. Newell, ``Dimensionality reduction for visualizing
  single-cell data using umap,'' \emph{Nature biotechnology}, vol.~37, no.~1,
  pp. 38--44, 2019.

\bibitem{ng2011sparse}
A.~Ng \emph{et~al.}, ``Sparse autoencoder,'' \emph{CS294A Lecture notes},
  vol.~72, no. 2011, pp. 1--19, 2011.

\bibitem{chen2020measuring}
D.~Chen, Y.~Lin, W.~Li, P.~Li, J.~Zhou, and X.~Sun, ``Measuring and relieving
  the over-smoothing problem for graph neural networks from the topological
  view,'' in \emph{Proceedings of the AAAI Conference on Artificial
  Intelligence}, vol.~34, no.~04, 2020, pp. 3438--3445.

\bibitem{bottou2010large}
L.~Bottou, ``Large-scale machine learning with stochastic gradient descent,''
  in \emph{Proceedings of COMPSTAT'2010}.\hskip 1em plus 0.5em minus
  0.4em\relax Springer, 2010, pp. 177--186.

\bibitem{srivastava2014dropout}
N.~Srivastava, G.~Hinton, A.~Krizhevsky, I.~Sutskever, and R.~Salakhutdinov,
  ``Dropout: a simple way to prevent neural networks from overfitting,''
  \emph{The journal of machine learning research}, vol.~15, no.~1, pp.
  1929--1958, 2014.

\bibitem{zhang2018generalized}
Z.~Zhang and M.~R. Sabuncu, ``Generalized cross entropy loss for training deep
  neural networks with noisy labels,'' in \emph{32nd Conference on Neural
  Information Processing Systems (NeurIPS)}, 2018.

\bibitem{yao2007early}
Y.~Yao, L.~Rosasco, and A.~Caponnetto, ``On early stopping in gradient descent
  learning,'' \emph{Constructive Approximation}, vol.~26, no.~2, pp. 289--315,
  2007.

\bibitem{zhu2010clustering}
L.~Zhu, B.~Ma, and X.~Zhao, ``Clustering validity analysis based on silhouette
  coefficient [j],'' \emph{Journal of Computer Applications}, vol.~30, no.~2,
  pp. 139--141, 2010.

\bibitem{ncir2021parallel}
C.-E.~B. Ncir, A.~Hamza, and W.~Bouaguel, ``Parallel and scalable dunn index
  for the validation of big data clusters,'' \emph{Parallel Computing}, vol.
  102, p. 102751, 2021.

\end{thebibliography}

\end{document}